\documentclass[conference]{IEEEtran}
\IEEEoverridecommandlockouts
\usepackage[utf8]{inputenc}

\usepackage{graphicx}
\usepackage{subcaption}
\usepackage{hyperref}
\usepackage{amsmath}
\usepackage{amsfonts}
\usepackage[version=4]{mhchem}
\usepackage{siunitx}
\usepackage{longtable,tabularx}
\usepackage{nicefrac}
\usepackage{rotating}
\usepackage{placeins}
\usepackage{tikz}
\newcommand{\vect}[1]{\boldsymbol{\mathbf{#1}}}
\def\realR{\mathbb{R}}
\def\sref{\operatorname{ref}}

\def\ie{i.e.}
\def\eg{e.g.}
\newcommand{\ccos}[1]{\operatorname{c}\!{#1}\;}
\newcommand{\ssin}[1]{\operatorname{s}\!{#1}\;}
\newcommand{\coss}[1]{\cos{#1}\;}
\newcommand{\sinn}[1]{\sin{#1}\;}

\newcommand{\norm}[1]{\left\lVert#1\right\rVert}

\def\xelem{\vect{i}_x^\top}
\def\yelem{\vect{i}_y^\top}
\def\zelem{\vect{i}_z^\top}
\def\atan2{\operatorname{atan2}}
\def\start{{\operatorname{start}}}
\def\send{{\operatorname{end}}}

\usepackage{xcolor}

\title{Aerobatic Trajectory Generation for a VTOL Fixed-Wing Aircraft Using Differential Flatness}

\author{Ezra Tal, Gilhyun Ryou, and Sertac Karaman
	\thanks{E. Tal, G. Ryou, and S. Karaman are with the Laboratory for Information and Decision Systems (LIDS), Massachusetts Institute of Technology. 
		{\tt\footnotesize \{eatal, ghryou, sertac\}@mit.edu}}%
}

\begin{document}

\maketitle

\begin{abstract}
This paper proposes a novel algorithm for aerobatic trajectory generation for a vertical take-off and landing (VTOL) tailsitter flying wing aircraft.
The algorithm differs from existing approaches for fixed-wing trajectory generation, as it considers a realistic six-degree-of-freedom (6DOF) flight dynamics model, including aerodynamics equations.
Using a global dynamics model enables the generation of aerobatics trajectories that exploit the entire flight envelope, enabling agile maneuvering through the stall regime, sideways uncoordinated flight, inverted flight etc.
The method uses the differential flatness property of the global tailsitter flying wing dynamics, which is derived in this work.
By performing snap minimization in the differentially flat output space, a computationally efficient algorithm, suitable for online motion planning, is obtained.
The algorithm is demonstrated in extensive flight experiments encompassing six aerobatics maneuvers, a time-optimal drone racing trajectory, and an airshow-like aerobatic sequence for three tailsitter aircraft.
\end{abstract}

\section*{Supplemental Material}
Video of the experiments can be found at \url{https://aera.mit.edu/projects/TailsitterAerobatics}.

\section{Introduction}
Vertical take-off and landing (VTOL) fixed-wing aircraft combine many of the advantages traditionally associated with either fixed-wing aircraft or rotorcraft.
They can exceed the range and endurance limitations typical of multicopters, while maintaining the capability to take-off, hover, and land in confined spaces.
This versatility is relevant to many real-world applications.
For example, transitioning search and rescue aircraft can cover large areas efficiently and closely inspect (indoor) areas of particular interest.
Similarly, VTOL delivery drones can safely make time-critical deliveries in remote environments without the need for a dedicated landing area.

Tailsitter VTOL aircraft transition between hover and forward flight by pitching, so that their rotors transition between lift generation and forward propulsion based on the attitude.
The tailsitter flying wing omits a tail and vertical surfaces, leading to a relatively simple mechanical design consisting of just a wing, two rotors, and two flaps that function as both elevators and ailerons.
By placing these flaps in the rotor wash and using differential thrust, the aircraft remains controllable throughout the flight envelope, including static conditions.
The simple, lightweight design allows a high thrust-to-weight ratio and the absence of a vertical tail surface reduces directional stability, leading to a highly agile and maneuverable aircraft.

\begin{figure}
	\centering
	\includegraphics[width=\linewidth]{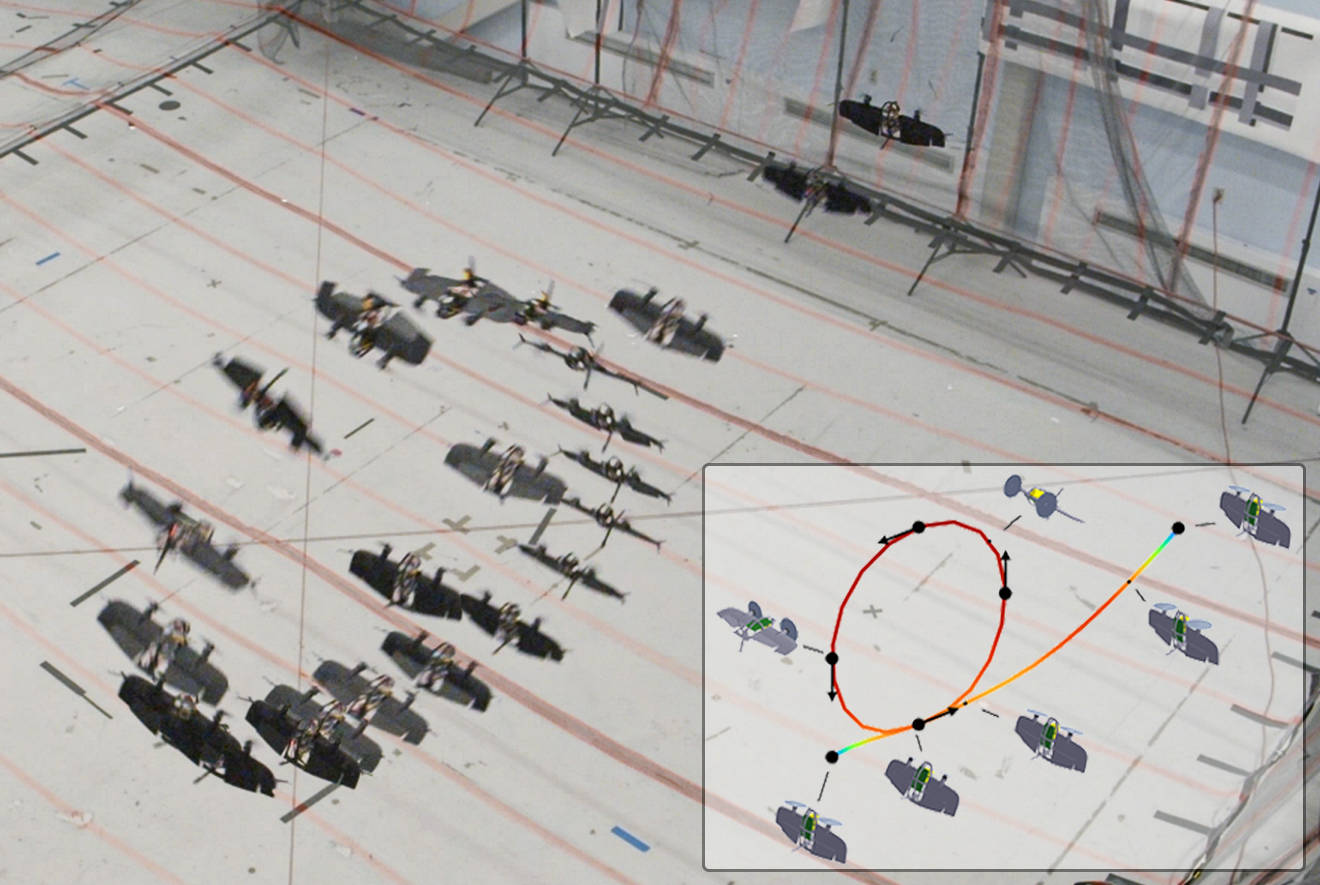}
	\caption{Loop trajectory reference and flight experiment.}
	\label{fig:loop_stack}
\end{figure}

In this paper, we show that, under some assumptions, the tailsitter flying wing flight dynamics are differentially flat.
This entails that the state and input variables can be expressed as a function of a flat output and a finite number of its derivatives~\cite{fliess1992lessystemesnon, fliessintro}.
Based on this flatness transform, we propose an algorithm for generating fast and agile tailsitter trajectories with low computational cost, \ie, suitable for online motion planning applications.
Our algorithm is capable of generating aerobatics maneuvers that exploit the entire flight envelope of the vehicle, including challenging conditions, such as sideways knife-edge flight and inverted flight, as shown in Fig. \ref{fig:loop_stack}.

Existing trajectory generation algorithms for fixed-wing aircraft often avoid the relatively complicated flight dynamics and instead use kinematics models.
For example, an extension of Dubins paths can be used to find the time-optimal trajectory with curvature constraints~\cite{chitsaz2007time}.
While accurate tracking of the resulting paths is not dynamically feasible due to the instantaneous acceleration changes needed to transition between straight lines and circular arcs, feedback control can be used to maintain a tracking error that is acceptable in calm flight~\cite{Owen2015}.
When considering fast and agile flight, the aircraft dynamics and control input constraints must be considered in trajectory generation, so that the resulting trajectory is dynamically feasible, \ie, so that it can be accurately tracked in flight.
Trajectory optimization subject to the six-degree-of-freedom (6DOF) nonlinear flight dynamics model is computationally costly, \eg,
optimization of the 4.5 m knife-edge maneuver presented by \cite{barry2014flying} takes 3--5 minutes of computation time (using direct collocation with twelve states and five control inputs) according to \cite{bry2015aggressive}.
Existing works address computational expense in various ways, \eg, by considering only (extended) point-mass equations of motion~\cite{van2009trajectory, morales2018equations, pashupathy2021unspecified}, by using a planner with pre-computed maneuvers~\cite{levin2019real, kim2020development}, by incorporating human-piloted expert demonstrations~\cite{cao2021real}, or by combining multiple simplified local dynamics models~\cite{beyer2021multi}.
In practice, these methods may impose limitations on the generated trajectories, especially when planning aerobatic maneuvers that rapidly progress through unconventional flight conditions.

In the context of trajectory generation, differential flatness enables transformation of trajectories from the flat output space to the state and control input space~\cite{fliess1992lessystemesnon,martin1992contribution}.
This property is widely leveraged towards computationally efficient trajectory generation and tracking for quadcopters by defining the trajectory in the flat output space consisting of the three-dimensional position and the yaw angle~\cite{mellinger2011minimum,richter2016polynomial,tal2020accurate}.
Differential flatness of fixed-wing aircraft dynamics has also been considered~\cite{martin1992contribution}.
However, the application of differential flatness towards trajectory generation for fixed-wing aircraft has mostly been limited to kinematics or simplified dynamics models.
Existing works consider path generation and tracking using a differentially flat coordinated flight model~\cite{hauser1997aggressive}
and aerobatics maneuvers using an aircraft kinematics model that does not incorporate angle of attack or sideslip angle~\cite{hall2008aerobatic}.
The algorithm presented in \cite{bry2015aggressive} is based on the differentially flat coordinated flight model given in \cite{hauser1997aggressive} and combines Dubins paths with a transverse polynomial offset to obtain smooth trajectories.

Our proposed method differs from existing flatness-based approaches for fixed-wing trajectory generation, as it considers a global 6DOF flight dynamics model, including aerodynamics equations.
By using a global dynamics model, we are able to generate aerobatics maneuvers that exploit the entire flight envelope, enabling agile maneuvering through the stall regime, sideways uncoordinated flight, inverted flight etc.
As we will show, the tailsitter flatness transform has a similar structure as the well-known quadcopter flat transform, in the sense that snap and yaw acceleration roughly correspond to the control inputs.
Hence, their reduction also increases feasibility of tailsitter trajectories, akin to the premise of minimum-snap trajectory generation algorithms for quadcopters~\cite{mellinger2011minimum,richter2016polynomial}.
This enables the application of similar efficient algorithms for minimum-snap trajectory generation in the flat output space towards generation of tailsitter aerobatics trajectories.

Our work contains several contributions.
Firstly, we propose an algorithm for aerobatic trajectory generation for a VTOL fixed-wing aircraft using differential flatness.
As far as we are aware, this is the first algorithm that uses differential flatness of a realistic fixed-wing flight dynamics model to generate aerobatic flight trajectories.
Secondly, we show differential flatness of the tailsitter flying wing dynamics model.
We note that recent work on trajectory-tracking flight control for a tailsitter flying wing also leverages differential flatness of the global dynamics model~\cite{tal2021global}.
However, this work does not include a method to obtain the control inputs as a function of the higher-order output derivatives, which is necessary for trajectory generation.
We present computational and experimental results that validate the suitability of the derived flatness transform to determine dynamic feasibility of candidate trajectories.
Thirdly, we provide extensive experimental results encompassing trajectories and flight tests for (i) six aerobatics maneuvers, (ii) a time-optimal drone racing trajectory at the limit of the vehicle's capability, and (iii) an airshow-like aerobatic sequence for three tailsitter aircraft.

The outline of this paper is as follows.
Section \ref{sec:prelims} presents preliminaries on the tailsitter flying wing flight dynamics and on minimum-snap trajectory generation.
The tailsitter flying wing flatness transform is derived in Section \ref{sec:diffflatness} and its suitability to predict dynamic feasibility of candidate trajectories is assessed in Section \ref{sec:dynfeas}.
Section \ref{sec:experiments} contains generated trajectories and experimental flight results for aggressive aerobatics maneuvers, a racing trajectory, and a multi-vehicle aerobatic sequence.
Finally, conclusions are given in Section \ref{sec:conclusion}.
\section{Preliminaries}
\label{sec:prelims}
\subsection{Flight Dynamics}\label{sec:flightdyn}
Our recent work on trajectory-tracking flight control~\cite{tal2021global} presented a global model of the tailsitter flying wing dynamics based on the $\varphi$-theory parameterization introduced by~\cite{lustosa2019global}.
In this section, we provide a brief overview of the dynamics model as a preliminary to the derivation of the corresponding flatness transform in Section \ref{sec:diffflatness}, which forms the basis of our trajectory generation algorithm.

\subsubsection{Vehicle Equations of Motion}\label{sec:eom}
\begin{figure}
	\centering
	\includegraphics[trim={12em 8em 12em 8em},clip,width=\linewidth]{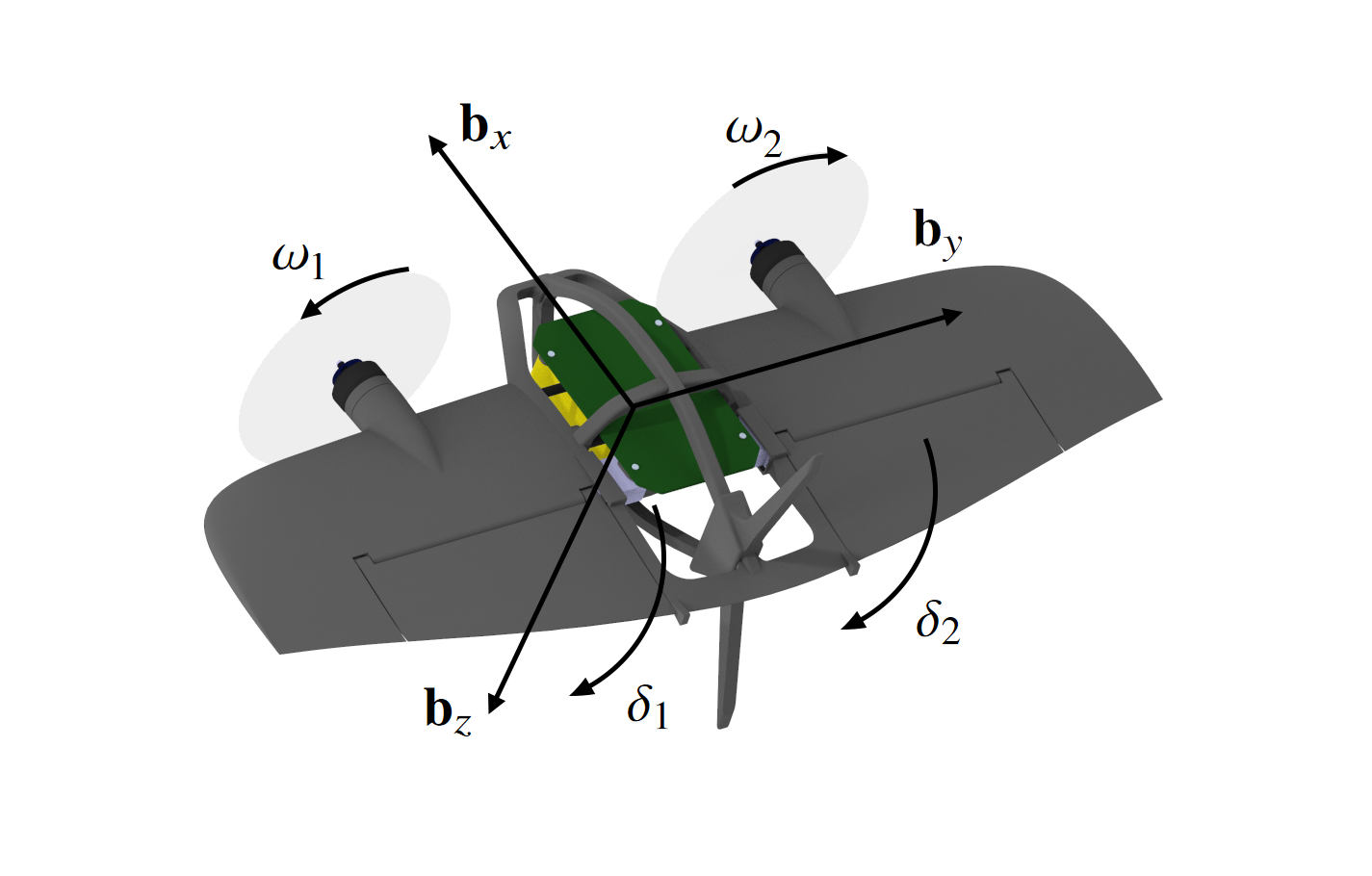}
	\caption{Body-fixed reference frame and control inputs.}
	\label{fig:bodyref_system}
\end{figure}

The vehicle translational dynamics are given by
\begin{align}
\dot{\vect{x}} &= \vect{v},\label{eq:xdot}\\
\dot{\vect{v}} &= g\vect{i}_z + m^{-1}\vect{R}^i_\alpha\vect{f}^\alpha ,\label{eq:vdot}
\end{align}
where $\vect{x}$ and $\vect{v}$ are respectively the vehicle position and velocity in the world-fixed reference frame, $g$ is the gravitational acceleration, and $m$ is the vehicle mass.
The vector $\vect{f}^\alpha$ represents the aerodynamic and thrust force in the vehicle-fixed zero-lift reference frame, and $\vect{R}^i_\alpha$ is the transformation matrix from this frame to the world-fixed reference frame, which is defined by the columns of the identity matrix $[\vect{i}_x\;\vect{i}_y\;\vect{i}_z]$.
The zero-lift frame differs from the general body-fixed reference frame, shown in Fig. \ref{fig:bodyref_system}, by a $-\alpha_0$ rotation around $\vect{b}_y$, where $\alpha_0$ is the zero-lift angle of attack.

The rotational dynamics are given by
\begin{align}
\dot{\vect{ \xi}} &= \frac{1}{2}\vect{\xi}\circ\vect{\Omega},\\
\dot{\vect{\Omega}} &= \vect{J}^{-1}(\vect{m} - \vect{\Omega} \times \vect{J}\vect{\Omega}), \label{eq:Omegadot}
\end{align}
where
$\vect{\Omega}$ is the angular velocity in the body-fixed reference frame, and
$\vect{\xi}$ is the normed quaternion attitude vector.
The matrix $\vect{J}$ is the vehicle moment of inertia tensor, and $\vect{m}$ represents the aerodynamic and thrust moment in the body-fixed reference frame.

\subsubsection{Force and Moment}\label{sec:aeromodel}
We employ $\varphi$-theory parameterization to obtain a global singularity-free model of the aerodynamic force and moment~\cite{lustosa2019global}.
The force in the zero-lift axis system is obtained by summing contributions due to thrust, flaps, and wings, as follows:
\begin{equation}\label{eq:force_total}
\vect{f}^\alpha = \vect{f}^\alpha_T + \vect{f}^\alpha_\delta +\vect{f}^\alpha_w.
\end{equation}
The thrust force is given by
\begin{equation}\label{eq:force_thrust}
\vect{f}^\alpha_T = \sum_{i=1}^2 \underbrace{\left[\begin{array}{c}
	\coss{\bar \alpha} (1 - c_{D_T})\\
	0\\
	\sinn{\bar \alpha}(c_{L_T} - 1)
	\end{array}\right]T_i}_{\vect{f}^\alpha_{T_i}},
\end{equation}
where $\bar \alpha$ is the sum of $\alpha_0$ and the thrust angle $\alpha_{T}$, $T_i$ is the thrust due to motor $i$, and the coefficients $c_{D_T}$ and $c_{L_T}$ represent drag and lift due to thrust vector components in the zero-lift axis system, respectively.
The motor thrust is computed as follows:
\begin{equation}\label{eq:thrust}
T_i = c_T \omega_i^2\;\;\;\text{with}\;\;i = 1, 2,
\end{equation}
where $c_T$ is the thrust coefficient and $\omega_i \geq 0$ is the speed of motor $i$.
The force contribution by the flaps is given by
\begin{equation}\label{eq:force_flap}
\vect{f}^\alpha_\delta = \sum_{i=1}^{2}\underbrace{-\left[\begin{array}{c}
	0\\
	0\\
	c^\delta_{L_T} \coss{\bar\alpha} T_i + c^\delta_{L_V}\|\vect{v}\| \xelem\vect{v}^\alpha
	\end{array}\right]\delta_i}_{\vect{f}^\alpha_{\delta_i}},
\end{equation}
where $\delta_i$ is the deflection angle of flap $i$.
Finally, the wing force contribution is obtained as
\begin{equation}\label{eq:force_body}
\vect{f}^\alpha_w = -\left[\begin{array}{c}
c_{D_V}\xelem \vect{v}^\alpha\\
0\\
c_{L_V}\zelem\vect{v}^\alpha
\end{array}\right]\|\vect{v}\|,
\end{equation}
where $c_{D_V}$ and $c_{L_V}$ are the wing drag and lift coefficients, respectively.
All aerodynamic coefficients incorporate air density, but could be scaled to account for attitude variations~\cite{tal2021algorithms}.
We note that \eqref{eq:force_total} does not contain any lateral force component, due to the absence of a fuselage and vertical tail surface.

The main moment contributions are due to the motors and flap deflections
\begin{equation}\label{eq:moment_total}
\vect{m} = \vect{m}_T + \vect{m}_\mu+\vect{m}_\delta.
\end{equation}
The moment due to motor thrust is given by
\begin{equation}
\vect{m}_T = \left[\begin{array}{c}
l_{T_y} \zelem \vect{R}^b_\alpha(\vect{f}^\alpha_{T_2} - \vect{f}^\alpha_{T_1})\\
c_{\mu_T} (T_1 + T_2)\\
l_{T_y} \xelem \vect{R}^b_\alpha(\vect{f}^\alpha_{T_1} - \vect{f}^\alpha_{T_2})
\end{array}\right],
\end{equation}
where $l_{T_y}$ is the moment arm and $c_{\mu_T}$ is the pitch moment coefficient due to thrust.
The moment due to motor torque is obtained as follows:
\begin{equation}
\vect{m}_\mu = \left[\begin{array}{c}
\cos\alpha_T\\
0\\
-\sin\alpha_T
\end{array}\right]\sum_{i=1}^2 \mu_i,
\end{equation}
where
\begin{equation}\label{eq:motortorque}
\mu_i = -(-1)^i c_\mu \omega_i^2\;\;\;\text{with}\;\;i = 1, 2,
\end{equation}
is the motor torque around the thrust-axis and $c_\mu$ is the propeller torque coefficient.
The flap contribution is given by
\begin{equation}\label{eq:moment_flap}
\vect{m}_\delta = \left[\begin{array}{c}
l_{\delta_y} \coss{\alpha_0}\zelem (\vect{f}^\alpha_{\delta_2} - \vect{f}^\alpha_{\delta_1})\\
l_{\delta_x} \zelem (\vect{f}^\alpha_{\delta_1} + \vect{f}^\alpha_{\delta_2})\\
l_{\delta_y} \sinn{\alpha_0}\zelem (\vect{f}^\alpha_{\delta_2} - \vect{f}^\alpha_{\delta_1})
\end{array}\right],
\end{equation}
where $l_{\delta_y}$ and $l_{\delta_x}$ are the relevant moment arms.
Moment contributions due to the freestream velocity and the angular velocity are neglected, as most of these are relatively small for the tailless flying wing and their inclusion may result in a much more complicated expression for the flatness transform.
An evaluation of the impact of modeling assumptions is provided in Section \ref{sec:dynfeas}.
\subsection{Minimum-Snap Trajectory Generation}\label{sec:tstaj_obj}
As we will show in Section \ref{sec:diffflatness}, the tailsitter dynamics model---with some simplifications---admits a differentially flat output
\begin{equation}\label{eq:traj}
\vect{\sigma}(t)=[\vect{x}(t)^\top\;\psi(t)]^\top,
\end{equation}
consisting of four elements: the vehicle position in the world-fixed reference frame $\vect{x}(t)\in \realR^3$, and the yaw angle $\psi(t)\in \mathbb{T}$, where $\mathbb{T}$ denotes the circle group.
Consequently, any sufficiently smooth output trajectory satisfies the dynamics \eqref{eq:xdot} through \eqref{eq:Omegadot} and, conversely, any state-space trajectory (including aerobatic trajectories with unconventional flight conditions) corresponds to a unique output trajectory \eqref{eq:traj}.
This bijective correspondence can be exploited to generate dynamically feasible aerobatics trajectories without resorting to computationally expensive state-space methods.

When focusing on aggressive flight trajectories, generation is complicated by the fact that the control input constraints, \ie, the motor speed and flap deflection limits, cannot readily be enforced in the flat output space.
Widely used algorithms for trajectory generation in the differentially flat output space of the quadcopter dynamics address this difficulty by minimizing snap, \ie, the fourth derivative of position, and yaw acceleration~\cite{mellinger2011minimum}.
In practice, this optimization roughly corresponds to reducing the required control moment and thus to increasing the likelihood that the control input limits are satisfied and the trajectory is feasible.
In Section \ref{sec:diffflatness}, we show that the flatness transform for the tailsitter dynamics has a similar form with control inputs depending on snap and yaw acceleration.
This makes snap minimization also suitable for generating aggressive tailsitter trajectories.

Elementary minimum-snap optimization subject to waypoint constraints can be formulated as follows:
\begin{equation}\label{eq:tstraj_minsnap}
\begin{aligned}
&\underset{\vect{\sigma}}{\text{minimize}}
& & \int_{0}^{T} \norm{\frac{d^4 \vect{x}}{dt^4}}^2 + \mu_\psi \Big(\frac{d^2 \psi}{dt^2}\Big)^2 dt\\
&  \text{subject to} & &  \vect{\sigma}\left(\sum \nolimits_{j=1}^{i}t_j\right) = \tilde{\vect{\sigma}}_i, \; i=0,\;\dots,\;m,\\
\end{aligned}
\end{equation}
where $\mu_\psi$ is a weighing parameter.
The nonnegative vector $\vect{t}$ represents the time allocation over the trajectory segments between the $m+1$ waypoints $\tilde{\vect{\sigma}}$ that must be attained in order.
Minimum-snap trajectory generation for quadcopters is widely studied, and various methods to obtain $\vect{t}$ have been proposed~\cite{mellinger2011minimum, richter2016polynomial, ryou2021multi}.
In principle, our framework for flatness-based trajectory generation is detached from the exact optimization formulation, enabling it to profit from the extensive research on minimum-snap trajectory generation, including extensions such as obstacle avoidance~\cite{deits2015efficient}.

In this paper, we use the formulation by \cite{richter2016polynomial} to describe the trajectory with piecewise polynomial functions that we define in terms of their derivatives at the waypoints.
For a given time allocation $\vect{t}$, the corresponding minimum-snap trajectory is then efficiently obtained in closed form using matrix multiplications, which we conveniently denote as
\begin{equation}
\vect{\sigma} = \vect{\chi}\left(\vect{t}, {\tilde{\boldsymbol{\sigma}}}, \dot{\tilde{\boldsymbol{\sigma}}}, \ddot{\tilde{\boldsymbol{\sigma}}}, \dots \right),
\end{equation}
where $\dot{\tilde{\boldsymbol{\sigma}}}$, $\ddot{\tilde{\boldsymbol{\sigma}}}$ etc. denote optional derivative constraints that may be set at some of the waypoints.
We first minimize snap subject to a rough estimate $\bar T$ of the total trajectory time based on the distance between waypoints, as follows:
\begin{equation}\label{eq:tstraj_minsnap2}
\begin{aligned}
&\underset{\vect{\sigma}, \vect{t}}{\text{minimize}}
& & \int_{0}^{\bar T} \norm{\frac{d^4 \vect{x}}{dt^4}}^2 + \mu_\psi \Big(\frac{d^2 \psi}{dt^2}\Big)^2 dt\\
&  \text{subject to} & &  \vect{\sigma} = \vect{\chi}\left(\vect{t}, {\tilde{\boldsymbol{\sigma}}}, \dot{\tilde{\boldsymbol{\sigma}}}, \ddot{\tilde{\boldsymbol{\sigma}}}, \dots \right),\\
&&& \sum \nolimits_{j=1}^{m}t_j = \bar T.
\end{aligned}
\end{equation}
In order to obtain aggressive aerobatic trajectories, we then minimize the scale factor $c$ that is applied to the resulting time allocation $\vect{t}$.
As such, we obtain the quickest minimum-snap trajectory $\vect{\sigma} = \vect{\chi}\left(c\vect{t}, {\tilde{\boldsymbol{\sigma}}}, \dot{\tilde{\boldsymbol{\sigma}}}, \ddot{\tilde{\boldsymbol{\sigma}}}, \dots \right)$ that is in the feasible set
\begin{equation}\label{eq:tstraj_feas1}
\Sigma_T = \Big\{\vect{\sigma}\Big|\vect{u}(t) \in \mathcal{U}\;\;\;\;
\forall t \in \left[0,T\right]\Big\},
\end{equation}
where $\vect{u}$ is the control input trajectory corresponding to $\vect{\sigma}$ and $\mathcal{U}$ is the set of permissible control inputs, \ie, the bounded set defined by the minimum and maximum allowed rotor speeds and flap deflections.
Additionally, we employ the method by \cite{ryou2021multi} to optimize the time allocation $\vect{t}$ using experimental evaluations, as described in Section \ref{sec:trajts_mfbo}.
\section{Differential Flatness Transform}\label{sec:diffflatness}
In recent work on tailsitter flight control, we have shown how the vehicle attitude and angular velocity can be obtained based on the trajectory \eqref{eq:traj} and its derivatives up to yaw rate and jerk (\ie, the third derivative of position)~\cite{tal2021global}.
In this section, we extend this derivation to obtain the full differential flatness transform, including an expression for the control inputs based on the trajectory derivatives up to yaw acceleration and snap.
This expression enables us to verify that the motor speeds and flap deflections corresponding to a candidate trajectory are permissible, \ie, that the trajectory is in the feasible set \eqref{eq:tstraj_feas1}.

\subsection{Attitude}\label{sec:diff_attthrust}
We first derive expressions for the attitude and collective thrust.
Rewriting \eqref{eq:vdot} as
\begin{equation}\label{eq:fi}
\vect{f}^i = m\left(\vect{a} - g\vect{i}_z\right)
\end{equation}
shows that the vehicle attitude and collective thrust are uniquely defined by three major constraints:
\begin{enumerate}
	\item[(i)] the yaw angle $\psi$,
	\item[(ii)] the fact that $\yelem \vect{f}^\alpha = 0$ according to \eqref{eq:force_thrust}, and
	\item[(iii)] the forces in the vehicle symmetry plane, \ie, $\xelem \vect{f}^\alpha$ and $\zelem\vect{f}^\alpha$.
\end{enumerate}
The Euler angles $\psi$, $\phi$, and $\theta$ in ZXY rotation sequence are used to describe the attitude.
These angles form a valid and universal attitude representation with each angle uniquely defined by one of the three constraints given above.
The angle symbols are also used to refer to rotation matrices between intermediate frames, \eg, the rotation matrix $\vect{R}^\phi_i$ represents the rotations by $\psi$ and $\phi$.

The yaw rotation $\psi \vect{i}_z$ is applied first and defines the direction of the horizontal component of $\vect{b}_y$.
Next, constraint (ii) is satisfied by the roll rotation
\begin{equation}\label{eq:roll}
\phi = -\atan2\left(\yelem \vect{R}^\psi_i \vect{f}^i,\zelem \vect{f}^i\right) + k\pi
\end{equation}
around the yawed $x$-axis $\vect{R}^i_\psi \vect{i}_x$,
where $\atan2$ is the four-quadrant inverse tangent function.
Constraint (ii) is satisfied $\forall k\in\{0,1\}$ and, in practice, $k$ can be set such that the obtained attitude trajectory is continuous.
Finally, constraint (iii) is satisfied by equating \eqref{eq:force_total} and \eqref{eq:fi} and solving for the collective thrust $T = T_1 + T_2$ and for the pitch rotation angle $\bar \theta$ from the frame $\phi$ to the zero-lift reference frame.

In solving these equations, we neglect the nonminimum phase dynamics due to the direct force contribution by the flaps.
When combined with feedback control, this approach achieves good trajectory generation and tracking performance for slightly nonminimum phase systems~\cite{hauser1992nonlinear}.
The method is simple and avoids the large and quickly changing control actions that exact feedback linearization of the nonminimum phase system may result in~\cite{tomlin1995output}.
We note that potentially a flat output of the nonminimum phase dynamics could be used to guarantee stable tracking~\cite{martin1996different}.
However, this approach requires defining the trajectory in terms of the center of oscillation instead of the vehicle center of mass, leading to difficulty with the relatively complicated 6DOF tailsitter dynamics model.

We substitute $\vect{f}^\alpha = \vect{R}^{\bar\theta}_\phi \vect{f}^\phi$ with $\vect{f}^\phi = \vect{R}^\phi_i \vect{f}^i$ as well as a similar expression for $\vect{v}^\alpha$ into \eqref{eq:force_total}
to obtain
\begin{multline}\label{eq:thetaTunsolved1}
\ccos{\bar\alpha} \left(1 - c_{D_T}\right) T - c_{D_V} \|\vect{v}\|\left(\ccos {\bar \theta} \xelem{\vect{v}^\phi} - \ssin {\bar \theta} \zelem \vect{v}^\phi \right) =\\ \ccos{{\bar \theta}}\xelem\vect{f}^\phi -\ssin{{\bar \theta}}\zelem{\vect{f}^\phi},
\end{multline}
\begin{multline}
\ssin{\bar\alpha} (c_{L_T} - 1) T - c_{L_V} \|\vect{v}\| \left(\ssin{{\bar \theta}}\xelem\vect{v}^\phi + \ccos{{\bar \theta}}\zelem{\vect{v}^\phi}\right) =\\ \ssin{{\bar \theta}}\xelem\vect{f}^\phi + \ccos{{\bar \theta}}\zelem\vect{f}^\phi,\label{eq:thetaTunsolved2} 
\end{multline}
where $\operatorname{c}$ and $\operatorname{s}$ represent cosine and sine, respectively.
Solving \eqref{eq:thetaTunsolved1} and \eqref{eq:thetaTunsolved2} for $\bar \theta$ and $T$ gives
\begin{multline}
{\bar \theta} = \atan2\\
\left(\eta \left(\xelem\vect{f}^\phi+c_{D_V}\|\vect{v}\|\xelem \vect{v}^\phi \right) -
 c_{L_V}\|\vect{v}\|\zelem\vect{v}^\phi-\zelem\vect{f}^\phi,\right.\\
\left.\eta \left(\zelem\vect{f}^\phi+c_{D_V}\|\vect{v}\|\zelem \vect{v}^\phi \right) + 
 c_{L_V}\|\vect{v}\|\xelem\vect{v}^\phi+\xelem\vect{f}^\phi\right) + k\pi,\label{eq:theta}
\end{multline}
\begin{multline}
T = \frac{1}{\ccos{\bar\alpha}\left(1-c_{D_T}\right)}\left(\ccos{{\bar \theta}}\xelem\vect{f}^\phi - \ssin{{\bar \theta}}\zelem\vect{f}^\phi +\right.\\
\left. c_{D_V} \|\vect{v}\|\left(\ccos {\bar \theta} \xelem{\vect{v}^\phi} - \ssin {\bar \theta} \zelem \vect{v}^\phi \right)\right),\label{eq:T}
\end{multline}
where
\begin{equation}
\eta = \frac{\ssin{\bar \alpha} \left(c_{L_T} - 1\right) }{\ccos{\bar\alpha}\left(1-c_{D_T}\right)}.
\end{equation}
is the ratio of lift and forward force due to thrust.
Again, the constraint is satisfied $\forall k\in\{0,1\}$ and, in practice, $k$ can be set such that the obtained attitude trajectory is continuous.
Finally, the pitch rotation of the body-fixed reference frame is obtained as $\theta = \bar\theta + \alpha_0$.

\subsection{Angular Velocity}\label{sec:diffangvel}
An expression for the angular velocity is obtained by taking the derivative of the Euler angles.
From \eqref{eq:roll}, we obtain
\begin{equation}\label{eq:phidot}
\dot \phi = -\frac{\dot\beta_x\beta_z-\beta_x\dot\beta_z}{\beta_x^2 + \beta_z^2},
\end{equation}
where $\beta_x$ and $\beta_z$ are respectively the first and second arguments of the $\atan2$ function, and
\begin{align}
\dot\beta_x &= -\ccos{\psi} \dot \psi \xelem\vect{f}^i - \ssin{\psi} \xelem\dot{\vect{f}}^i - \ssin{\psi}\dot\psi\yelem\vect{f}^i + \ccos{\psi}\yelem\dot{\vect{f}}^i,\label{eq:betadot}\\
\dot\beta_z &= \zelem\dot{\vect{f}}^i,\label{eq:beta2dot}
\end{align}
with, from the derivative of \eqref{eq:fi},
\begin{equation}\label{eq:forceder}
\dot{\vect{f}}^i = m\vect{j}.
\end{equation}
Similarly, from \eqref{eq:theta} we obtain
\begin{equation}\label{eq:dottheta}
\dot \theta = \frac{\dot\sigma_x\sigma_z - \sigma_x\dot\sigma_z }{\sigma_x^2 + \sigma_z^2},
\end{equation}
where $\sigma_x$ and $\sigma_z$ are the respective arguments of the $\atan2$ function, and
\begin{align}
\dot\sigma_x &= \eta \left(\xelem\dot{\vect{f}}^\phi + c_{D_V} \tau_x \right) - c_{L_V}\tau_z - \zelem\dot{\vect{f}}^\phi,\label{eq:sigma1dot}\\
\dot\sigma_z &= \eta \left(\zelem\dot{\vect{f}}^\phi + c_{D_V} \tau_z \right) + c_{L_V}\tau_x + \xelem\dot{\vect{f}}^\phi\label{eq:sigma2dot}
\end{align}
with
\begin{align}
\tau_x &= \dot{\|\vect{v}\|} \xelem \vect{v}^\phi + \|\vect{v}\|\xelem\dot{\vect{v}}^\phi,\\
\tau_z &= \dot{\|\vect{v}\|} \zelem \vect{v}^\phi + \|\vect{v}\|\zelem\dot{\vect{v}}^\phi
\end{align}
and
\begin{align}
\dot{\|\vect{v}\|} &= \frac{\vect{v}^\top\vect{a}}{\|\vect{v}\|},\\
\dot{\vect{v}}^\phi &= \dot{\vect{R}}^\phi_i\vect{v} + \vect{R}^\phi_i\vect{a}.\label{eq:accel}
\end{align}
The expression for the force derivative $\dot{\vect{f}}^\phi$ is similar to \eqref{eq:accel}.
As described in Section \ref{sec:diff_attthrust}, we neglect the direct force contribution by the flaps.
Finally, we obtain the angular velocity in the body-fixed reference frame, as follows:
\begin{equation}\label{eq:angrate}
\vect{\Omega} = \left[\begin{array}{c}
0\\\dot \theta\\0
\end{array}\right] + \vect{R}^{ \theta}_\phi \left[\begin{array}{c}
\dot \phi \\0\\0
\end{array}\right] + \vect{R}^{ \theta}_\psi \left[\begin{array}{c}
0\\0\\\dot \psi
\end{array}\right].
\end{equation}

\subsection{Motor Speeds and Flap Deflections}\label{sec:diffcontrol}
In order to obtain the control inputs, we first derive an expression for the angular acceleration as a function of snap and yaw acceleration.
By taking the derivative of \eqref{eq:phidot}, we obtain the following expression for the roll acceleration
\begin{multline}\label{eq:ddotphi}
\ddot{\phi} = \left(\beta_x^2+\beta_z^2\right)^{-2}\left(\left(\dot\beta_x\beta_z-\beta_x\dot\beta_z\right)\left(2\beta_x\dot\beta_x+2\beta_z\dot\beta_z\right)-\right.\\
\left.\left(\ddot\beta_x\beta_z-\beta_x\ddot\beta_z\right)\left(\beta_x^2+\beta_z^2\right)\right),
\end{multline}
where
\begin{align}
\begin{split}
\ddot \beta_x &= \left(\ssin{\psi}{\dot{\psi}}^2 - \ccos{\psi}\ddot\psi\right)\xelem\vect{f}^i - 2 \ccos{\psi} \dot\psi \xelem \dot{\vect{f}}^i - \ssin{\psi}\xelem \ddot{\vect{f}}^i\\
&\;\;\;\;\;- \left(\ccos{\psi}{\dot \psi}^2 + \ssin{\psi}\ddot \psi\right)\yelem \vect{f}^i -2 \ssin{\psi}\dot\psi\yelem\dot{\vect{f}}^i +\ccos{\psi}\yelem\ddot{\vect{f}}^i,
\end{split}
\\
\ddot \beta_z &= \zelem \ddot{\vect{f}}^i
\end{align}
are obtained as the derivatives of \eqref{eq:betadot} and \eqref{eq:beta2dot}, and
the second force derivative is a function of snap
\begin{equation}
\ddot{\vect{f}}^i = m\vect{s}.
\end{equation}
Similarly, by taking the derivative of \eqref{eq:dottheta} we obtain the pitch acceleration
\begin{multline}\label{eq:ddottheta}
\ddot \theta = \big(\left(\ddot\sigma_x\sigma_z-\sigma_x\ddot\sigma_z\right)\left(\sigma_x^2+\sigma_z^2\right) -\\
\left(\dot\sigma_x\sigma_z-\sigma_x\dot\sigma_z\right)\left(2\sigma_x\dot\sigma_x+2\sigma_z\dot\sigma_z\right)\big){\left(\sigma_x^2+\sigma_z^2\right)^{-2}},
\end{multline}
where
\begin{align}
\ddot\sigma_x &= \eta \left(\xelem\ddot{\vect{f}}^\phi + c_{D_V} \dot{\tau}_x \right) - c_{L_V}\dot{\tau}_z - \zelem\ddot{\vect{f}}^\phi,\label{eq:sigma1ddot}\\
\ddot\sigma_z &= \eta \left(\zelem\ddot{\vect{f}}^\phi + c_{D_V} \dot{\tau}_z \right) + c_{L_V}\dot{\tau}_x + \xelem\ddot{\vect{f}}^\phi\label{eq:sigma2ddot}
\end{align}
with
\begin{align}
\dot{\tau}_x &= \ddot{\|\vect{v}\|} \xelem \vect{v}^\phi + 2\dot{\|\vect{v}\|} \xelem \dot{\vect{v}}^\phi + \|\vect{v}\|\xelem\ddot{\vect{v}}^\phi,\\
\dot{\tau}_z &= \ddot{\|\vect{v}\|} \zelem \vect{v}^\phi + 2\dot{\|\vect{v}\|} \zelem \dot{\vect{v}}^\phi + \|\vect{v}\|\zelem\ddot{\vect{v}}^\phi
\end{align}
and
\begin{align}
\ddot{\|\vect{v}\|} &= \frac{\vect{a}^\top\vect{a} + \vect{v}^\top\vect{j}}{\|\vect{v}\|} - \frac{\vect{v}^\top\vect{a} \dot{\|\vect{v}\|}}{\|\vect{v}\|^2},\\
\ddot{\vect{v}}^\phi &= \ddot{\vect{R}}^\phi_i\vect{v} + 2\dot{\vect{R}}^\phi_i\vect{a} + \vect{R}^\phi_i\vect{j}.\label{eq:jerk}
\end{align}
The expression for the force second derivative $\ddot{\vect{f}}^\phi$ is similar to \eqref{eq:jerk}.
We combine the roll acceleration and pitch acceleration obtained from respectively \eqref{eq:ddotphi} and \eqref{eq:ddottheta} with the yaw acceleration $\ddot{\psi}$ to obtain the angular acceleration in the body-fixed reference frame.
We take the derivative of \eqref{eq:angrate} to obtain the following expression:
\begin{multline}\label{eq:angacc}
\dot{\vect{\Omega}} = \left[\begin{array}{c}
0\\\ddot \theta\\0
\end{array}\right] + \dot{\vect{R}}^{ \theta}_\phi \left[\begin{array}{c}
\dot \phi \\0\\0
\end{array}\right] + \vect{R}^{ \theta}_\phi \left[\begin{array}{c}
\ddot \phi \\0\\0
\end{array}\right] +\\ \dot{\vect{R}}^{ \theta}_\psi \left[\begin{array}{c}
0\\0\\\dot \psi
\end{array}\right] + \vect{R}^{ \theta}_\psi \left[\begin{array}{c}
0\\0\\\ddot \psi
\end{array}\right].
\end{multline}

We can now find the moment in the body-fixed reference frame by rewriting \eqref{eq:Omegadot}, as follows:
\begin{equation}\label{eq:moment}
\vect{m} = \vect{J}\dot{\vect{\Omega}} + \vect{\Omega}\times \vect{J}\vect{\Omega}.
\end{equation}
Next, we solve \eqref{eq:moment_total} for the flap deflections and differential thrust $\Delta T = T_1 - T_2$.
We find an expression for $\Delta T$ by equating 
\begin{equation}
\zelem\left( \vect{m}_T + \vect{m}_\mu\right) = \zelem \vect{m},
\end{equation}
which assumes that the contribution by $\zelem \vect{m}_\delta$ is negligible.
Due to the multiplication with $\sin{\alpha_0}$, this assumption typically does not result in significant discrepancies.
Using $\mu_1 + \mu_2 = \nicefrac{c_\mu}{c_T}\Delta T$, we obtain
\begin{multline}\label{eq:DeltaT}
\Delta T = {\zelem \vect{m}}\big(- \ssin{\alpha_T}\frac{c_\mu}{c_T} +\\ 
	l_{T_y}\left(\ccos{\alpha_0}\ccos{\bar \alpha}(1 - c_{D_T}) - \ssin{\alpha_0}\ssin{\bar\alpha}(c_{L_T}- 1)\right) \big)^{-1}.
\end{multline}
The individual thrust values are then given by
\begin{equation}\label{eq:T1T2}
T_1 = \frac{T + \Delta T}{2},\;\;\;\;\;\;\;\;\;\;\;\;
T_2 = \frac{T - \Delta T}{2},
\end{equation}
and the motor speeds can be obtained from \eqref{eq:thrust}.
For the flap deflections, we deduct $\vect{m}_T$ and $\vect{m}_\mu$ from $\vect{m}$ to obtain $\vect{m}_\delta$, and we
rewrite \eqref{eq:moment_flap}, as follows:
\begin{equation}\label{eq:d1d2}
\left[\begin{array}{c}
\delta_1\\
\delta_2
\end{array}\right] = \left[\begin{array}{cc}
-l_{\delta_y} \ccos{\alpha_0} {\nu}_1& l_{\delta_y} \ccos{\alpha_0} {\nu}_2\\
l_{\delta_x}  {\nu}_1&l_{\delta_x}  {\nu}_2
\end{array}\right]^{-1}\left[\begin{array}{c}
\xelem \vect{m}_\delta\\
\yelem \vect{m}_\delta
\end{array}\right]
\end{equation}
with
\begin{equation}
{\nu}_i = -c^\delta_{L_T} \coss{\bar\alpha} T_i - c^\delta_{L_V}\|\vect{v}\| \xelem\vect{v}^\alpha.
\end{equation}
Note that---since the control inputs cannot instantaneously change---dynamic feasibility of $\vect{\sigma}$ requires continuity of \eqref{eq:moment}, and therefore at least fourth-order continuity of the position $\vect{x}$ and at least second-order continuity of the yaw $\psi$.
\section{Dynamic Feasibility}\label{sec:dynfeas}
In this section, we evaluate the suitability of the flat transform presented in Section \ref{sec:diffflatness} to determine feasibility of a candidate trajectory on the actual vehicle.
Specifically, the transform is used to obtain the control inputs that are then evaluated according to \eqref{eq:tstraj_feas1}.
For a description of the vehicle parameters and their estimation, the reader is referred to \cite{tal2021algorithms}.

\subsection{Hover-to-Hover Trajectory}

\begin{figure}
	\centering
	\includegraphics[trim={40em 0 40em 0},clip,width=\linewidth]{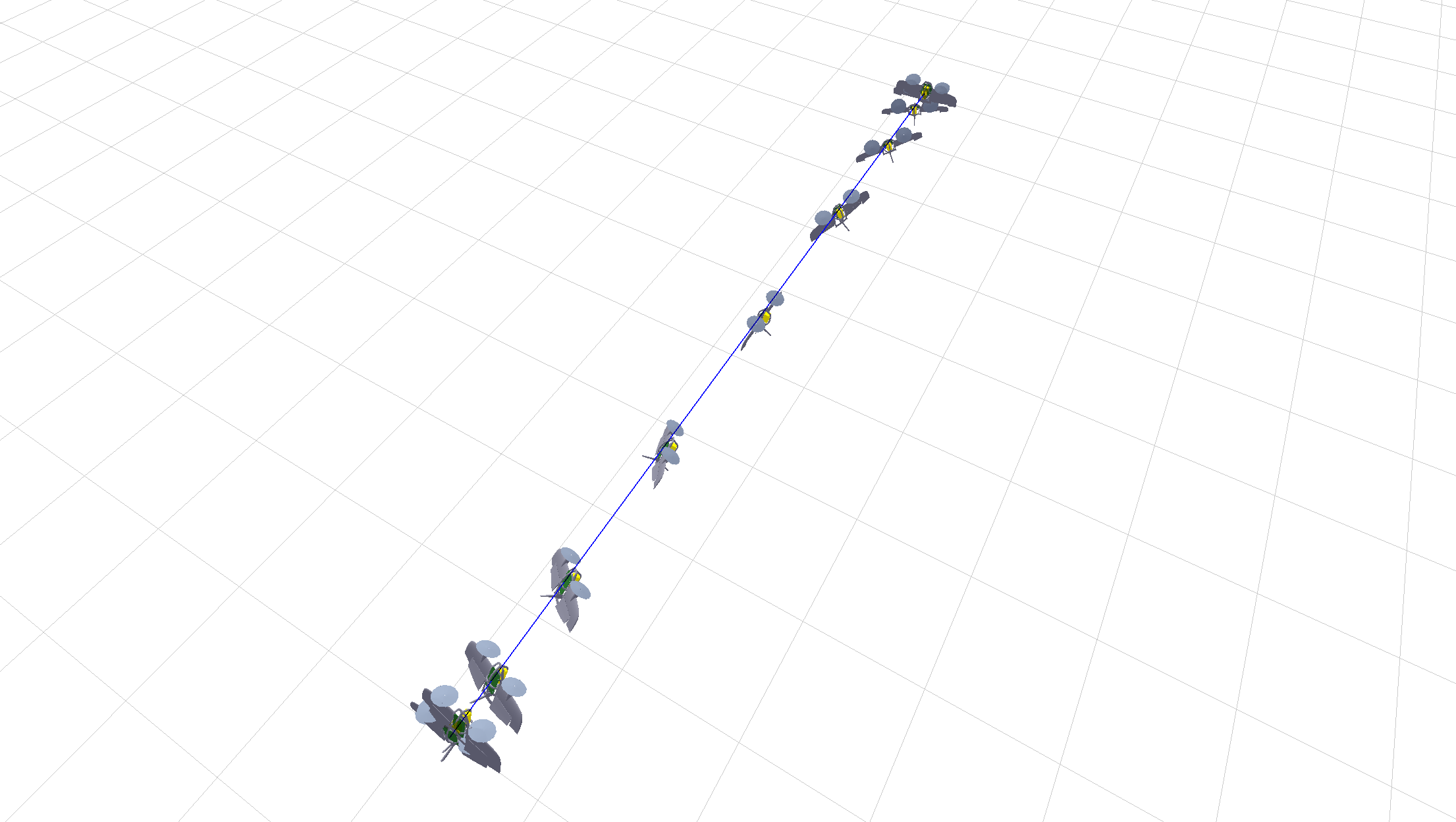}
	\caption{6 m hover-to-hover trajectory with $\psi^{\start} = 0$ rad, $\psi^{\send} = \pi$ rad. Trajectory time is 3 s, interval between poses is 0.25 s.}
	\label{fig:p2p_example}
\end{figure}

\begin{figure}
		\centering
		\includegraphics[width=\linewidth]{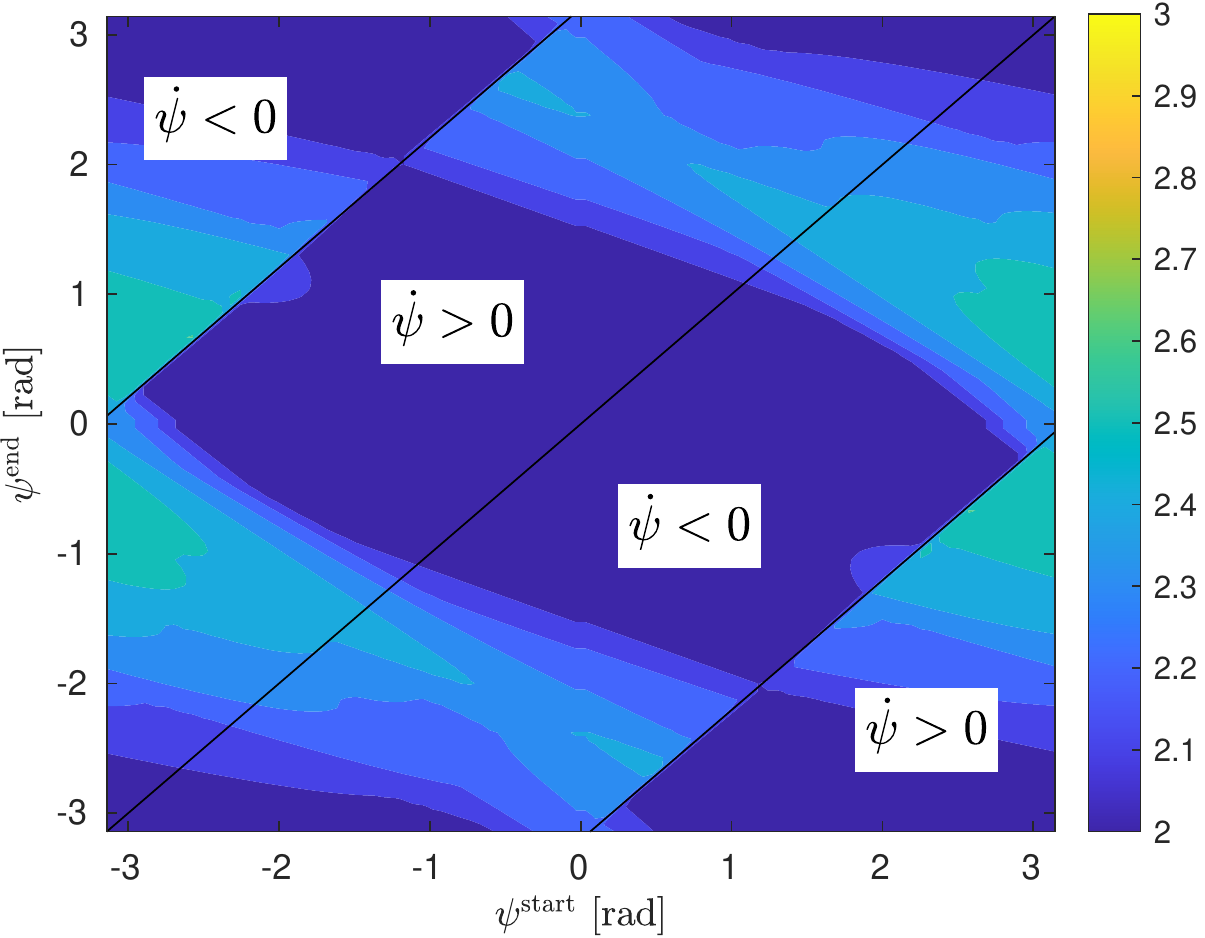}
		\caption{Minimum feasible time for 6 m hover-to-hover trajectory using minimal yaw rotation.}
		\label{fig:p2p_feas_swr}
\end{figure}

\begin{figure*}
	\centering
	
	\begin{subfigure}[t]{0.485\textwidth}
		\centering
		\includegraphics[width=\linewidth]{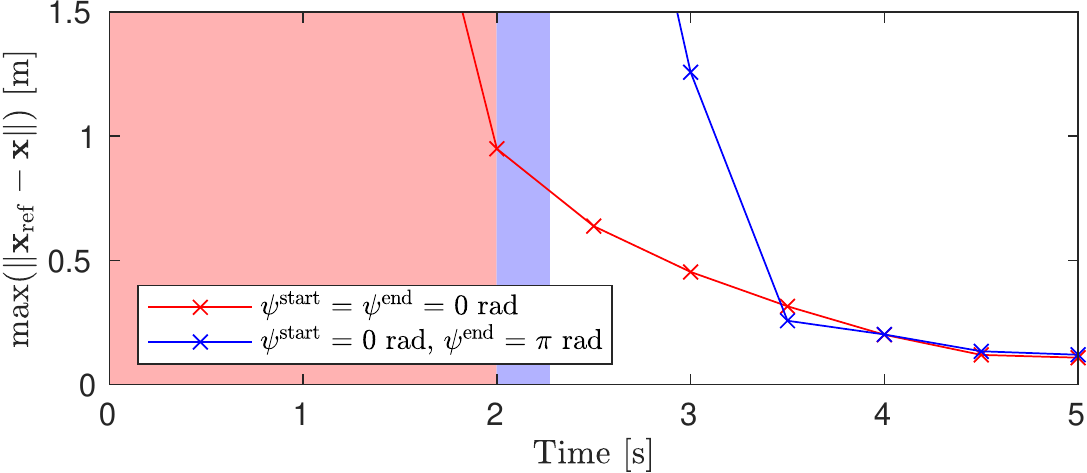}
		\caption{Maximum position tracking error with and without yaw rotation.}
		\label{fig:p2p_exp_0-314}
	\end{subfigure}%
	\quad
	\begin{subfigure}[t]{0.485\textwidth}
		\centering
		\includegraphics[width=\linewidth]{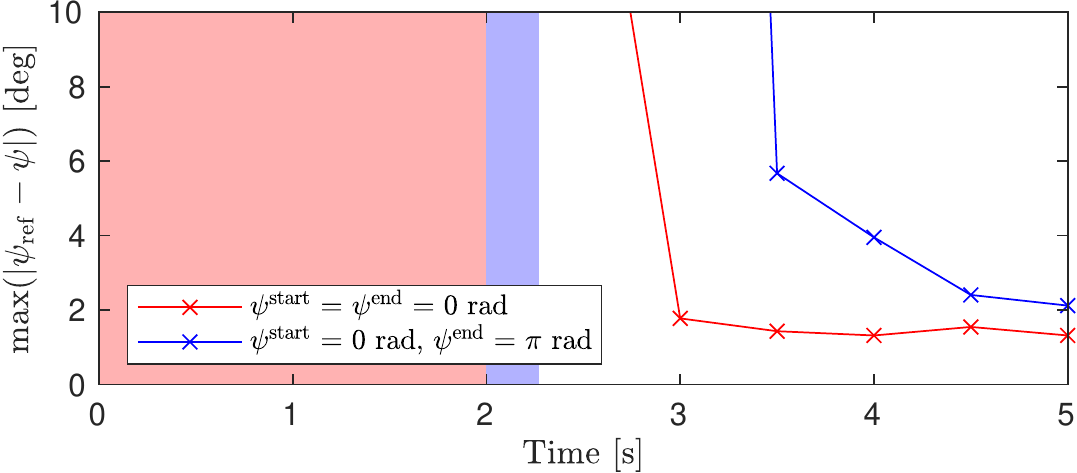}
		\caption{Maximum yaw tracking error with and without yaw rotation.}
		\label{fig:p2p_exp_0-314_yaw}
	\end{subfigure}%
	
	\caption{Tracking error in flight experiments 6 m hover-to-hover trajectory. Shaded area indicates infeasibility according to differential flatness transform.}
	\label{fig:p2p_exp}
\end{figure*}

We consider a single-segment hover-to-hover trajectory with
\begin{align}
	\tilde{\vect{\sigma}}_0 &= \left[\begin{array}{cccc}
	0 \;\;\;\;\;\;\; & 0& 0& \psi^\start\\
	\end{array}\right]^\top,\\
	\tilde{\vect{\sigma}}_1&= \left[\begin{array}{cccc}
	6 \;\text{[m]} & 0& 0& \psi^\send\\
	\end{array}\right]^\top.
\end{align}
This trajectory requires large acceleration and simultaneous yawing motion through the transition regime.
Based on the flatness transform described in Section \ref{sec:diffflatness}, we determine the minimal feasible time for the minimum-snap trajectory with various $\psi^\start$ and $\psi^\send$.
An example trajectory is shown in Fig. \ref{fig:p2p_example}.
Figure \ref{fig:p2p_feas_swr} shows results for the trajectory with yawing motion from $\psi^\start$ to $\psi^\send$ using the minimal rotation.
It can be seen that the fastest times are achieved in the center of the figure, around  $\psi^\start = \psi^\send = 0$ rad, which corresponds to forward coordinated flight.
We observe discontinuity along the yaw direction switching lines, which indicates that, in some cases, it may be beneficial to yaw in the opposite direction.
However, in practice the difference is typically small, meaning that the minimal rotation that is obtained from optimization in the flat output space is (nearly) optimal.

We conduct experiments to compare the feasibility boundary from Fig. \ref{fig:p2p_feas_swr} to the tracking error of the actual vehicle.
Figure \ref{fig:p2p_exp} shows the tracking error for the hover-to-hover trajectory in coordinated flight without yaw, \ie, $\psi^\start = \psi^\send = 0$ rad, and for the same trajectory but with $\psi^\start = 0$ rad, $\psi^\send = \pi$ rad.
Each point on the curves corresponds to a flight experiment.
As the trajectory time on the horizontal axis increases, the maneuvers become less aggressive, and the tracking error decreases.
The corresponding feasibility boundaries predicted in Fig. \ref{fig:p2p_feas_swr} are indicated by the colored shading, \ie, the shaded areas in the left of the figure correspond to infeasible trajectory times.
While only a single color is shown at a time, the infeasibility areas continue from their boundary all the way to the vertical axis on the left.
For the yawing trajectory, the tracking error increases at lower speeds compared to the coordinated flight trajectory, as predicted by the feasibility boundaries.
We note that these boundaries correspond to the most aggressive trajectories that theoretically can be tracked by the given vehicle dynamics model, neglecting practical factors such as modeling errors and imperfect state estimation and control, so that it is expected that significant tracking error occurs before they are reached.
The coordinated flight trajectory at the feasibility boundary (2.0 s) attains a maximum speed of 7.6 m/s within 1 s and attains a maximum load of 3.1$g$. It is tracked with less than 1 m position tracking error.

\subsection{Circular Trajectory}

\begin{figure*}
	\centering
	\begin{subfigure}[t]{0.32\textwidth}
		\centering
		\includegraphics[width=\linewidth]{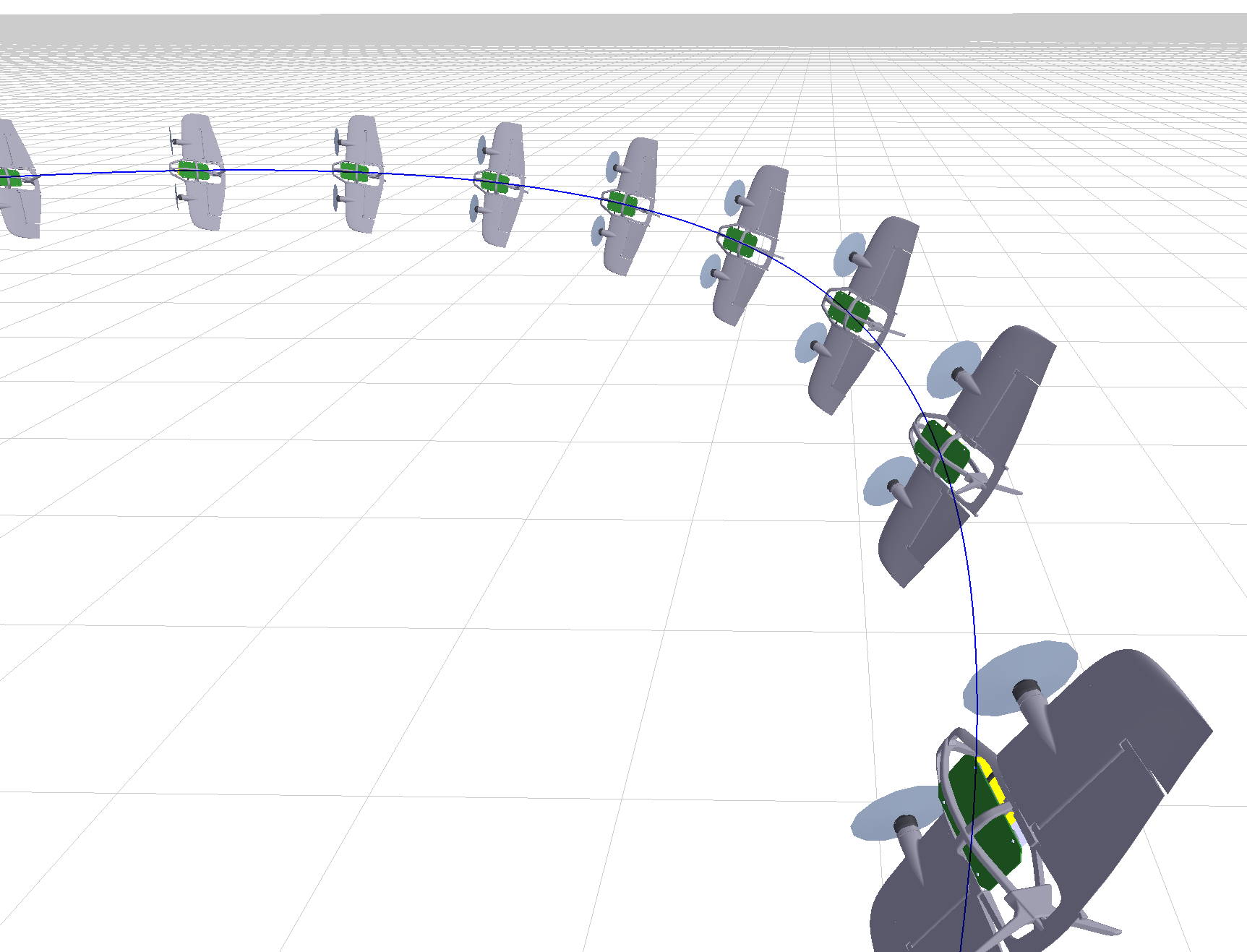}
		\caption{Coordinated.}
		\label{fig:circle_ref_coor}
	\end{subfigure}%
	~
	\begin{subfigure}[t]{0.32\textwidth}
		\centering
		\includegraphics[width=\linewidth]{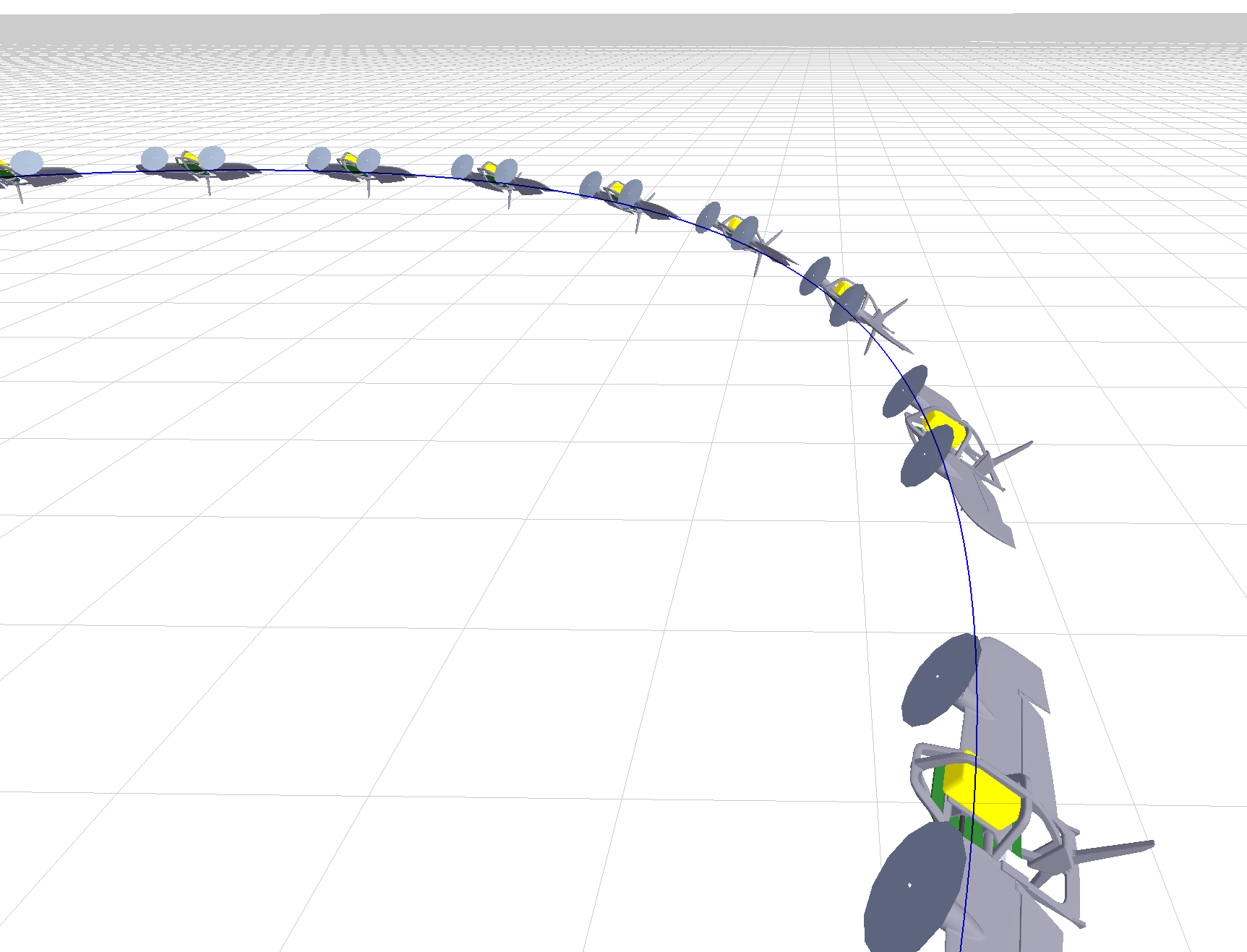}
		\caption{Knife edge.}
		\label{fig:circle_ref_ke}
	\end{subfigure}%
	~
	\begin{subfigure}[t]{0.32\textwidth}
		\centering
		\includegraphics[width=\linewidth]{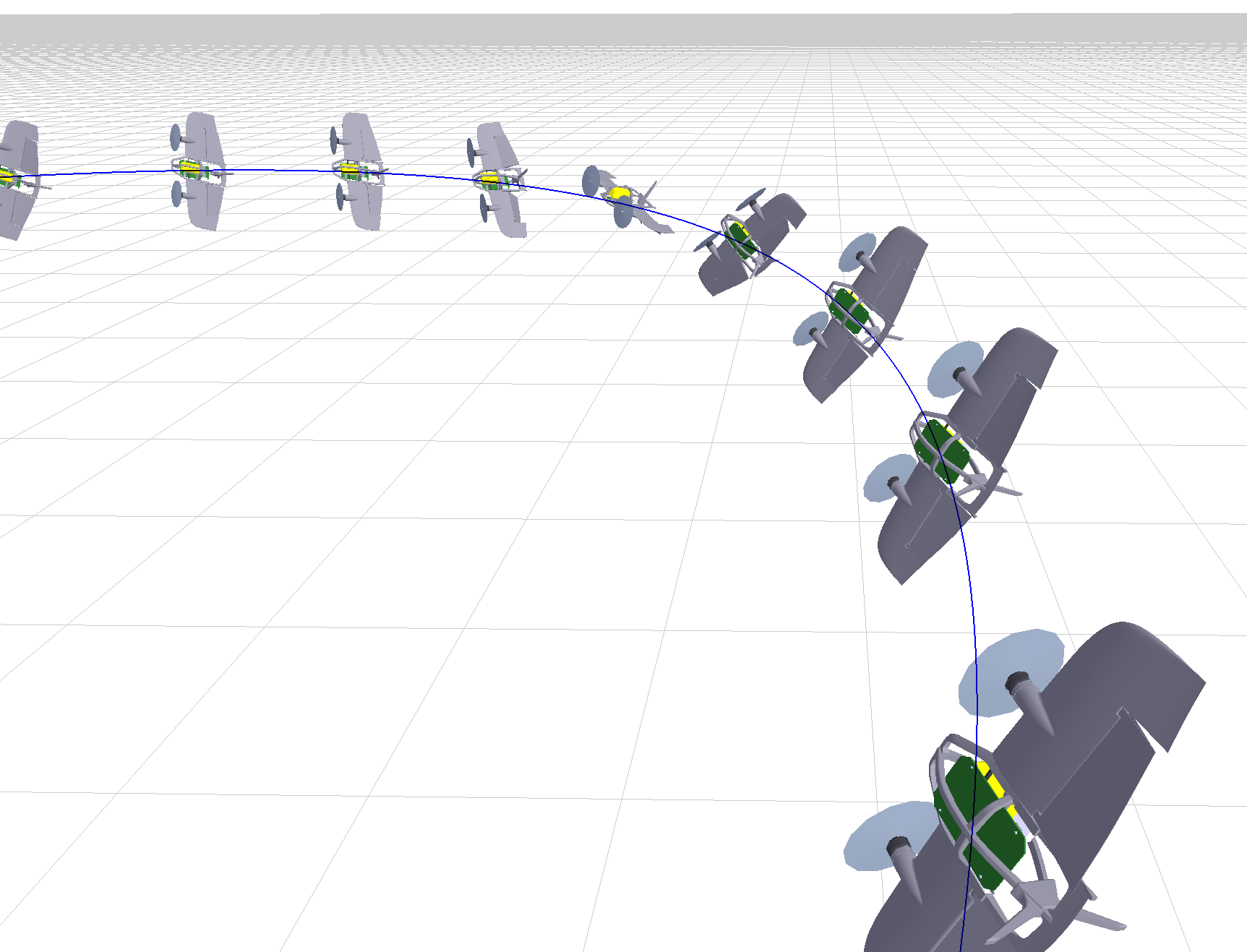}
		\caption{Rolling.}
		\label{fig:circle_ref_rc}
	\end{subfigure}%
	\caption{Circular trajectory with various yaw references.}
	\label{fig:circle_ref}
\end{figure*}

\begin{figure*}
	\centering
	
	\begin{subfigure}[t]{0.485\textwidth}
		\centering
		\includegraphics[width=\linewidth]{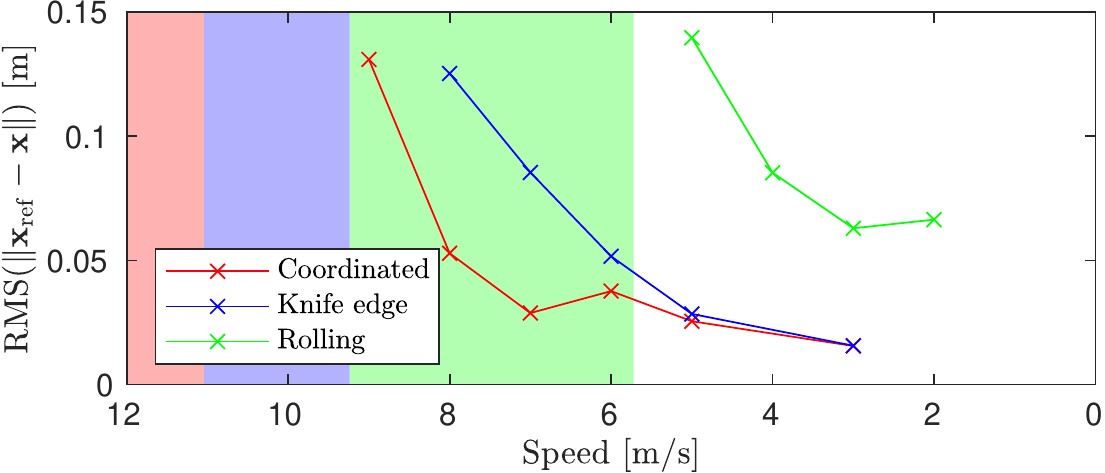}
		\caption{RMS position tracking error.}
		\label{fig:circle_exp_pos}
	\end{subfigure}%
	\quad
	\begin{subfigure}[t]{0.485\textwidth}
		\centering
		\includegraphics[width=\linewidth]{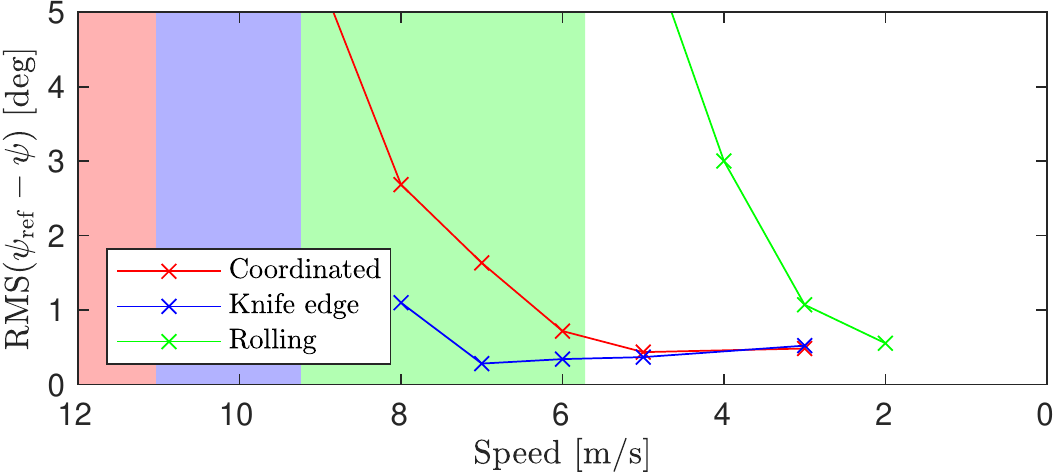}
		\caption{RMS yaw tracking error.}
		\label{fig:circle_exp_yaw}
	\end{subfigure}%
	\caption{Tracking error in flight experiments for circular trajectory with various yaw references. Shaded area indicates infeasibility according to differential flatness transform.}
	\label{fig:circle_exp}
\end{figure*}

\begin{table}
	\centering
	\footnotesize
	\setlength{\tabcolsep}{5pt}
	\setlength{\extrarowheight}{2pt}
	\begin{subtable}{\linewidth}
		\centering
		\begin{tabular}{ccc|ccc|ccc|ccc}
			\hline
			\multicolumn{3}{c}{Velocity}&
			\multicolumn{3}{c}{Acceleration}&
			\multicolumn{3}{c}{Jerk}&
			\multicolumn{3}{c}{Snap}\\
			\hline
			$\vect{i}_x$ & $\vect{i}_y$ & $\vect{i}_z$&
			$\vect{i}_x$ & $\vect{i}_y$ & $\vect{i}_z$&
			$\vect{i}_x$ & $\vect{i}_y$ & $\vect{i}_z$&
			$\vect{i}_x$ & $\vect{i}_y$ & $\vect{i}_z$\\
			$v$ & 0 & 0 &
			0 & $-\Omega v$ & 0 &
			$-\Omega^2 v$ & 0 & 0&
			0 & $\Omega^3 v$ & 0\\
			\hline
		\end{tabular}
		\caption{Position derivatives.}
		\label{tab:circ_ref_pos}
	\end{subtable}
	
	\begin{subtable}{\linewidth}
		\centering
		\begin{tabular}{ccc|ccc|ccc}
			\hline
			\multicolumn{3}{c}{Coordinated}&
			\multicolumn{3}{c}{Knife edge}&
			\multicolumn{3}{c}{Rolling}\\
			\hline
			$\psi$ & $\dot \psi$ & $\ddot \psi$&
			$\psi$ & $\dot \psi$ & $\ddot \psi$&
			$\psi$ & $\dot \psi$ & $\ddot \psi$\\
			0 & $-\Omega$ & 0 &
			$\nicefrac{\pi}{2}$ & $-\Omega$ & 0&
			$[0,2\pi]$& $\Omega$&0\\
			\hline
		\end{tabular}
		\caption{Yaw (derivatives).}
		\label{tab:circ_ref_yaw}
	\end{subtable}
	
	\caption{Flat output (derivatives) for various circular trajectories with $\Omega = \nicefrac{v}{r}$.}
	\label{tab:circ_ref}
\end{table}

In order to evaluate the accuracy of the feasibility prediction at high speed and large sustained acceleration, we use the flatness transform to determine the maximum speed on a circular trajectory with a 3 m radius.
As shown in Fig. \ref{fig:circle_ref}, we consider two trimmed conditions, coordinated and knife-edge flight, as well as a rolling/yawing motion where $\psi$ changes at the same rate but in the opposite direction.
The position and yaw derivatives for evaluation of the feasibility are given in Table \ref{tab:circ_ref}.

We perform experiments for all three circular trajectories at various speeds.
The results are shown in Fig. \ref{fig:circle_exp}, where each point on the curves corresponds to a flight experiment.
The figure is oriented similarly to Fig. \ref{fig:p2p_exp} with the most aggressive, \ie, the highest speed, trajectories towards the left.
It shows that the flat dynamics model predicts that coordinated flight can be performed up to the highest speed, followed by knife-edge flight, and finally the rolling circle, which has a relatively low maximum speed.
The position tracking errors obtained from flight experiments agree with this prediction.
Figure \ref{fig:circle_exp_pos} shows the expected increase in each position tracking error before the corresponding shaded area is reached.
Similar behavior can be observed in Fig. \ref{fig:circle_exp_yaw} for the yaw tracking error on the coordinated and rolling circle.
The yaw error in knife-edge flight remains very small, even at high speeds, because---in this condition---the vehicle orientation reduces the sensitivity of yaw to attitude errors and increases the yaw control effectiveness by differential thrust.

Since the flat transform does not consider lateral forces on the tailless aircraft, the speed in circular knife-edge flight is mostly limited by the maximum thrust.
In fact, completely neglecting the aerodynamics and solving for the maximum speed
\begin{equation}\label{eq:vmax_circleke}
v_{\max} = \sqrt{2 c_T \bar \omega ^2\frac{  r}{m}}
\end{equation}
with $\bar \omega$ the maximum motor speed,
results in only a small overestimation when compared to the maximum speed obtained from the flat transform (9.5 m/s versus 9.2 m/s).
In flight experiments, the vehicle achieved RMS position and yaw tracking errors of respectively 12.5 cm and 1.1 deg at 8 m/s, approaching the theoretical maximum speed with relatively small tracking error.
Considering that at least some control input margin must be maintained to enable stabilization of the unstable knife-edge condition (making the theoretical limit unattainable), this affirms that the lateral aerodynamic force must indeed be quite small and can be neglected in the flat dynamics model.

Considering the comparative results for both trajectories, we can conclude that the differential flatness transform gives a useful prediction of the critical trajectory time or speed where we can expect to observe a stark increase in tracking error on the real vehicle.

\section{Flight Experiments}\label{sec:experiments}
\begin{figure}
	\centering
	\includegraphics[trim={10em 20em 5em 20em},clip,width=\linewidth]{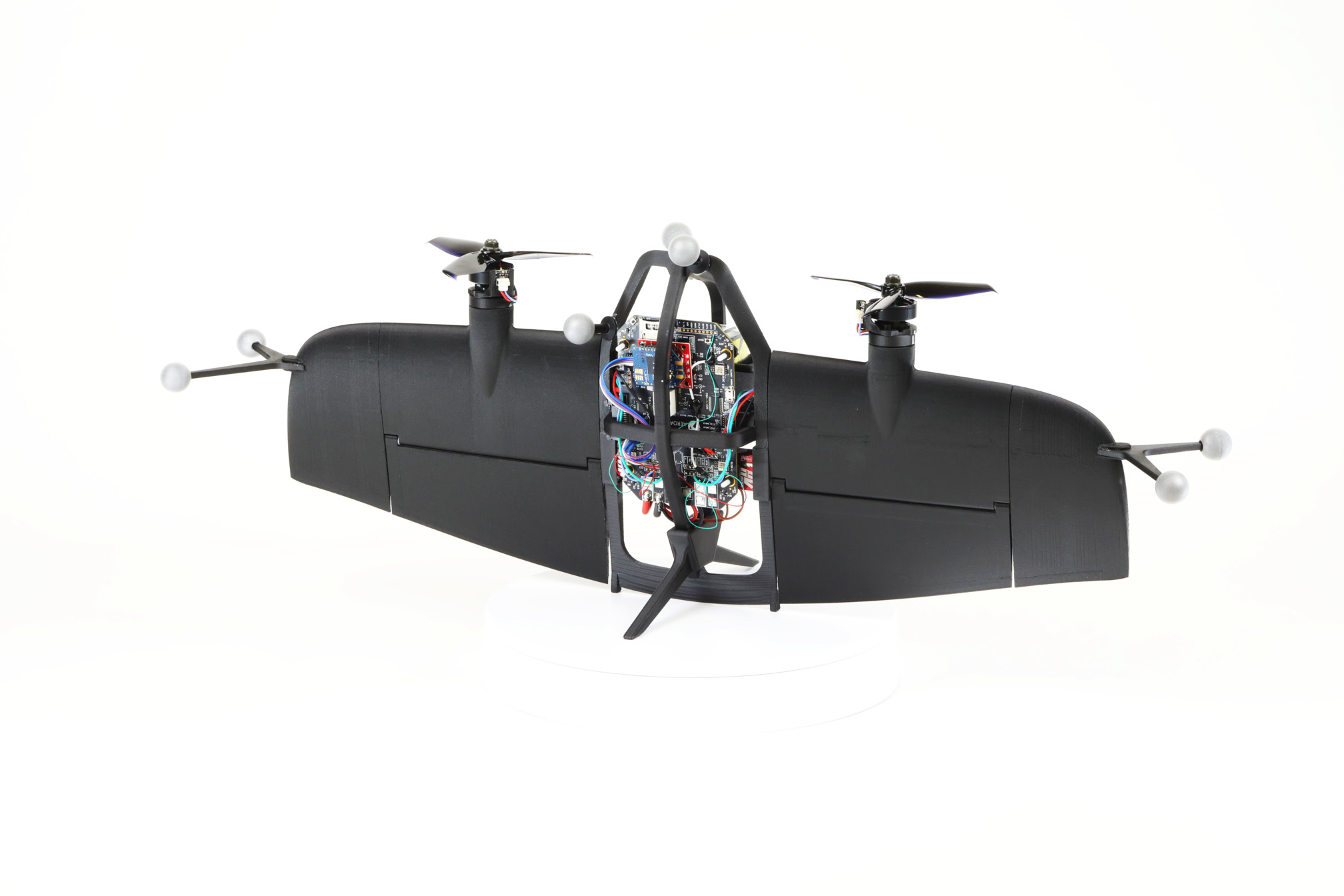}
	\caption{Tailsitter flying wing aircraft used in the experiments.}
	\label{fig:aircraft}
\end{figure}

We present extensive experimental results to validate the generated aerobatic trajectories.
These flight tests demonstrate six types of aerobatic maneuvers, a racing trajectory through a sequence of gates, and an airshow-like aerobatic sequence with three tailsitter aircraft that aggressively maneuver in close proximity to obstacles and to each other.
Detailed descriptions of the experimental platform, shown in Fig. \ref{fig:aircraft}, and the trajectory-tracking flight control system are given in \cite{bronz2020mission} and \cite{tal2021global}, respectively.
A video of the experiments can be found at \url{https://aera.mit.edu/projects/TailsitterAerobatics}.

\subsection{Aerobatic Maneuvers}\label{sec:aerobatic}
\begin{table*}
	\centering
	\caption{Maximum speed, load, and angular rate for reference trajectories; and maximum and root mean square (RMS) position tracking error for flight experiments.}
	\label{tab:control_exp_error}
	\begin{tabular}{lccccc}
		\hline
		& max$\|\vect{v}\|$ [m/s]&max$\|\vect{a} - \vect{i}_z g\|$ [$g$]& max$\|\vect{\Omega}\|$ [deg/s] & max$\|\vect{x}_{\sref}-\vect{x}\|$ [m] & RMS$\|\vect{x}_{\sref}-\vect{x}\|$ [m]\\
		\hline
		Loop                    & 3.8 & 2.4 & 665 & 0.74 & 0.39\\
		Knife-Edge Flight       & 5.0 & 1.1 & 376 & 0.85 & 0.33\\
		Climbing Turn           & 5.3 & 3.1 & 647 & 0.97 & 0.53\\
		Immelmann Turn          & 6.0 & 2.1 & 538 & 0.94 & 0.43\\
		Split S                 & 5.0 & 1.4 & 415 & 0.63 & 0.25\\
		Differential Thrust Turn& 8.0 & 1.6 & 312 & 1.37 & 0.63\\
		\hline
	\end{tabular}
\end{table*}

We first demonstrate how the flatness transform enables generation of aerobatic maneuvers using relatively simple waypoint (derivative) constraints in the trajectory output space.
The resulting maneuvers exploit the full flight envelope of the tailsitter aircraft, including post-stall and sideways flight, and do not require restrictive assumptions such as coordinated flight and curvature limitations.
The data shown in Table \ref{tab:control_exp_error} confirms the aerobatic character of the trajectories, which reach speeds up to 8.0 m/s, loads of over 3$g$, and angular rates that exceed 650 deg/s.
Conforming to the observations from Section \ref{sec:dynfeas}, we found that each maneuver can be slowed down to reduce tracking error but we chose to accept some tracking error in favor of increased aggressiveness.

\subsubsection{Loop}
\begin{figure*}
	\centering
	
	\begin{subfigure}[t]{0.5\textwidth}
		\centering
		\includegraphics[trim={0em 20em 0em 10em},clip,width=\linewidth]{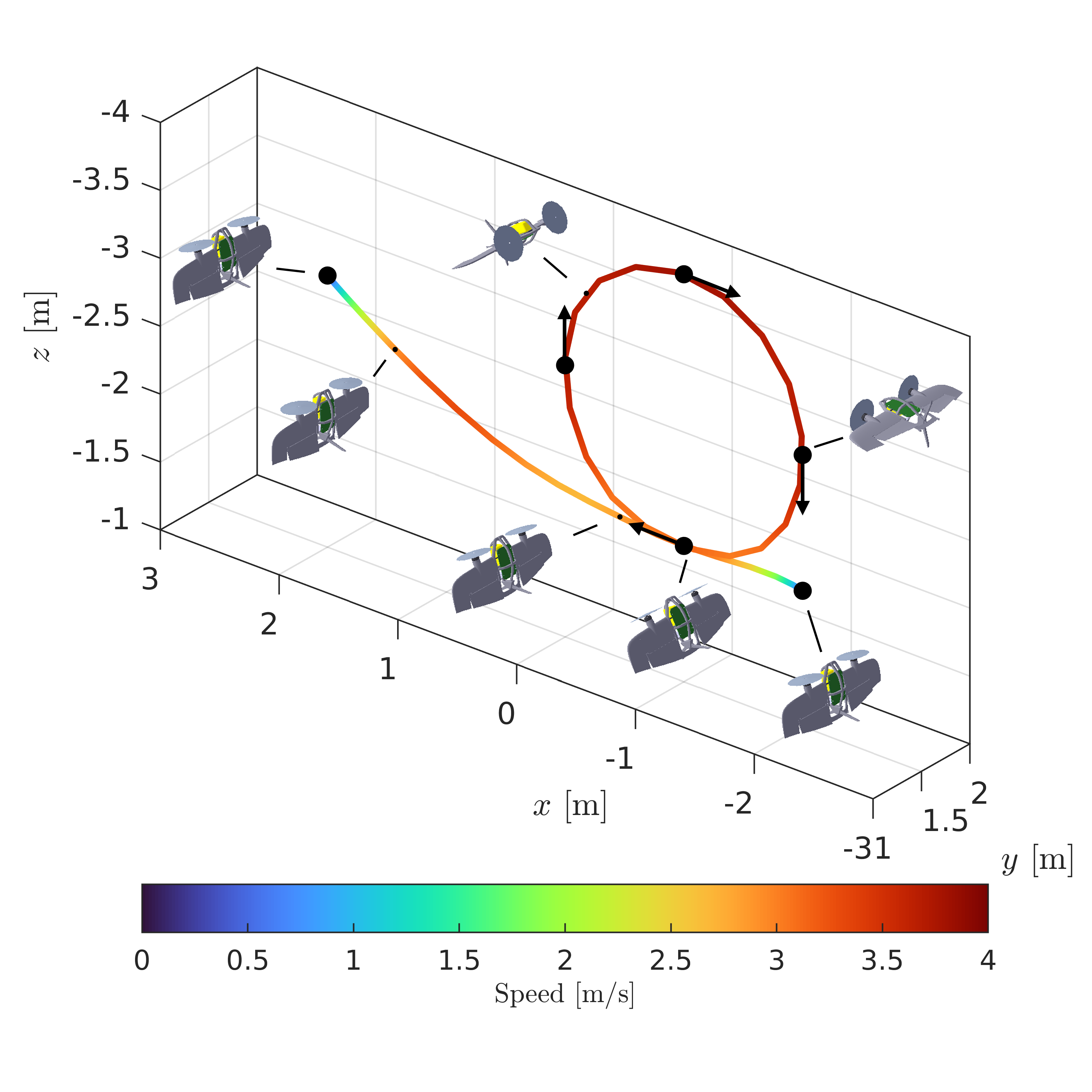}
		\caption{Reference with waypoints. Start and end points are static hover, and arrows indicate velocity direction constraints.}
		\label{fig:loop_ref}
	\end{subfigure}%
	\begin{subfigure}[t]{0.5\textwidth}
		\centering
		\includegraphics[trim={0em 20em 0em 10em},clip,width=\linewidth]{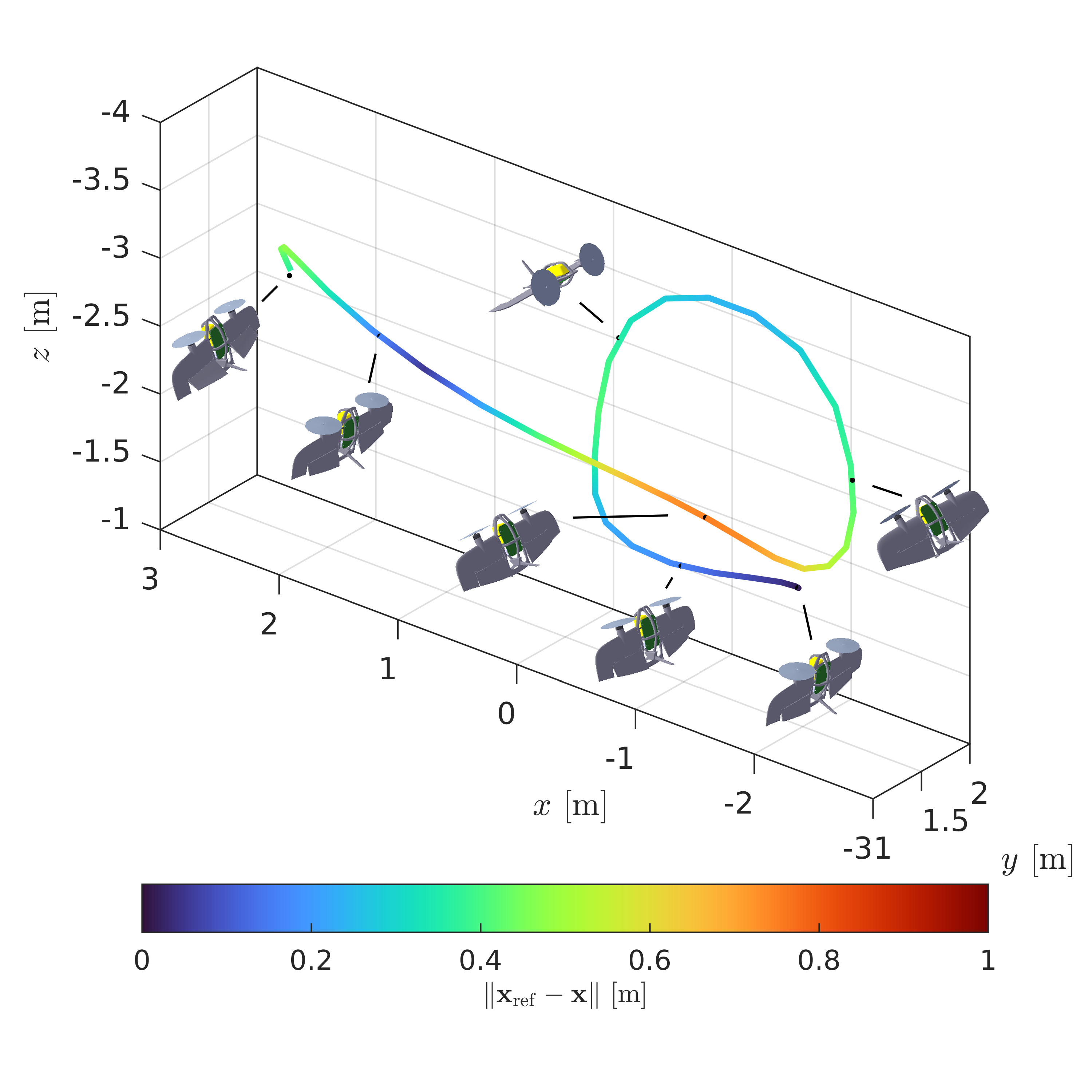}
		\caption{Experiment.}
		\label{fig:loop_exp}
	\end{subfigure}%
	\caption{Loop. Interval between poses is 0.7 s.}
	\label{fig:loop}
\end{figure*}

\begin{figure}
	\centering
		\includegraphics[width=\linewidth]{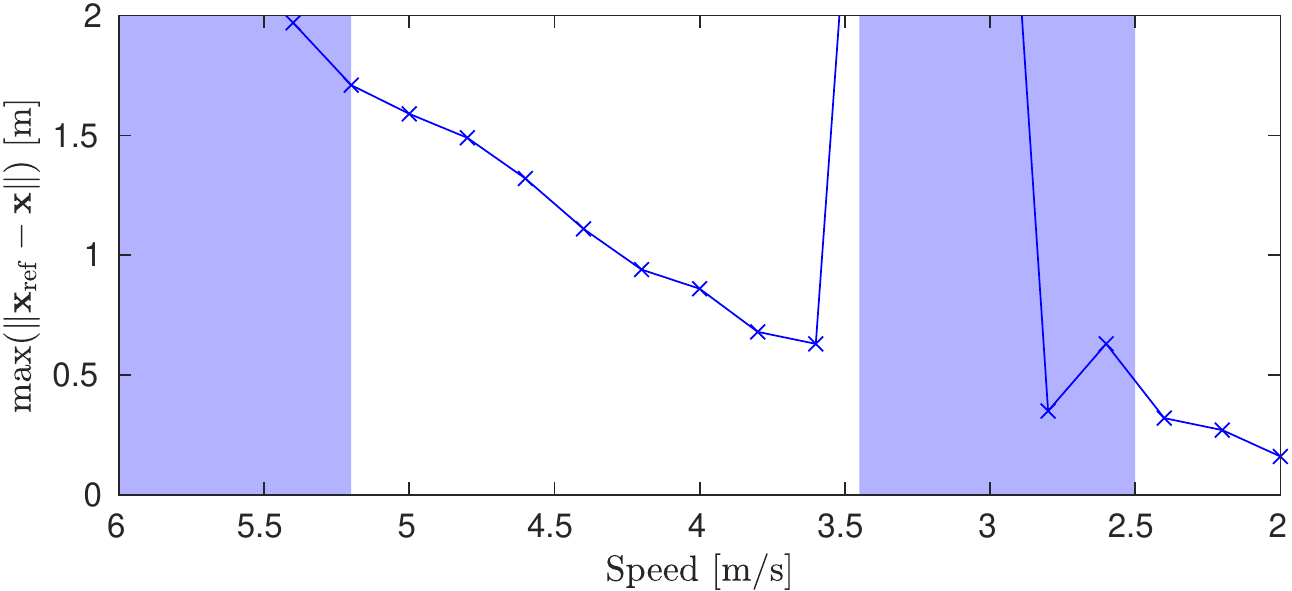}
	\caption{Maximum position tracking error in flight experiments for loop trajectory at various speeds. Shaded areas indicate infeasibility according to differential flatness transform and coincide with increased experimental tracking error.}
	\label{fig:loop_feas}
\end{figure}

The loop trajectory shown in Fig. \ref{fig:loop} consists of five waypoints (of which two coincide) on a vertical circle with 1 m radius, and start and end points constrained to static hover.
We add tangential velocity constraints to enforce a circular path.
As shown in \mbox{Fig. \ref{fig:loop_feas}}, the loop trajectory has several feasibility boundaries. When flown slowly (\ie, below 2.5 m/s), the trajectory is feasible and flown in hover attitude with $\theta \approx \nicefrac{\pi}{2}$ rad.
When flown faster (\ie, around 4.5 m/s), the vehicle performs a loop, making a full upward pitch rotation.
Intermediate speeds (\ie, around 3 m/s) are too slow to perform a loop and require the vehicle to quickly pitch back down at the top of the circular segment, rendering the trajectory infeasible due to flap deflection limits.
The maximum position tracking error obtained from flight experiments shows a stark increase in this region of infeasibility and also increases as the infeasibility boundary at very high speed (\ie, 5.2 m/s) is approached.
The trajectory with a maximum speed of 3.8 m/s is shown in Fig. \ref{fig:loop}.
The loop maneuver is successfully performed in the flight experiment. The maximum position error of 74 cm is incurred when exiting the final circular segment.

\subsubsection{Knife-Edge Flight}

\begin{figure*}
	\centering
	
	\begin{subfigure}[t]{0.5\textwidth}
		\centering
		\includegraphics[trim={10em 0em 30em 0},clip,width=\linewidth]{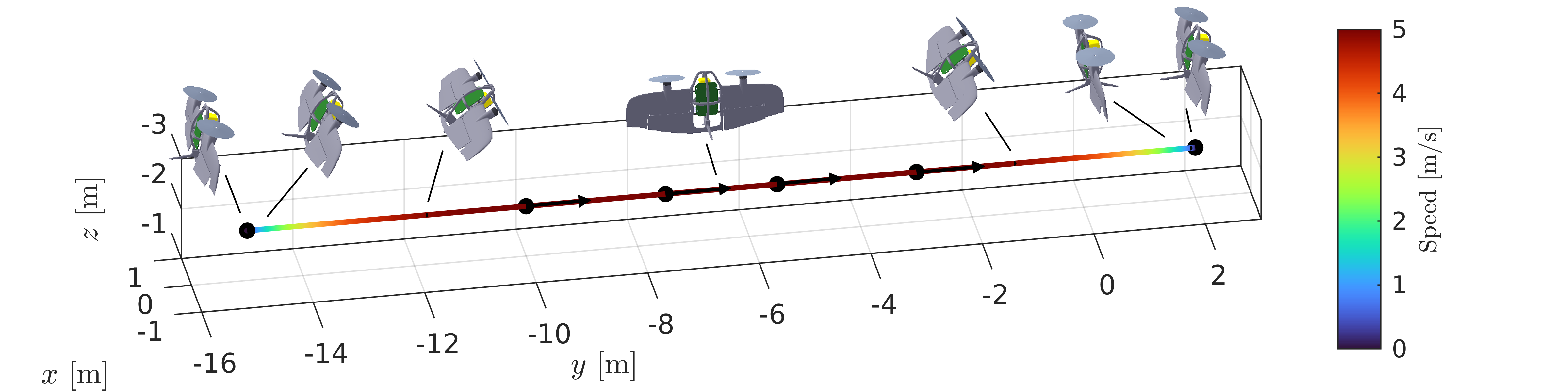}
		\caption{Reference with waypoints. Start and end points are static hover, and arrows indicate 5 m/s velocity constraints.}
		\label{fig:knifeedgetraj_ref}
	\end{subfigure}%
	\begin{subfigure}[t]{0.5\textwidth}
		\centering
		\includegraphics[trim={10em 0em 30em 0},clip,width=\linewidth]{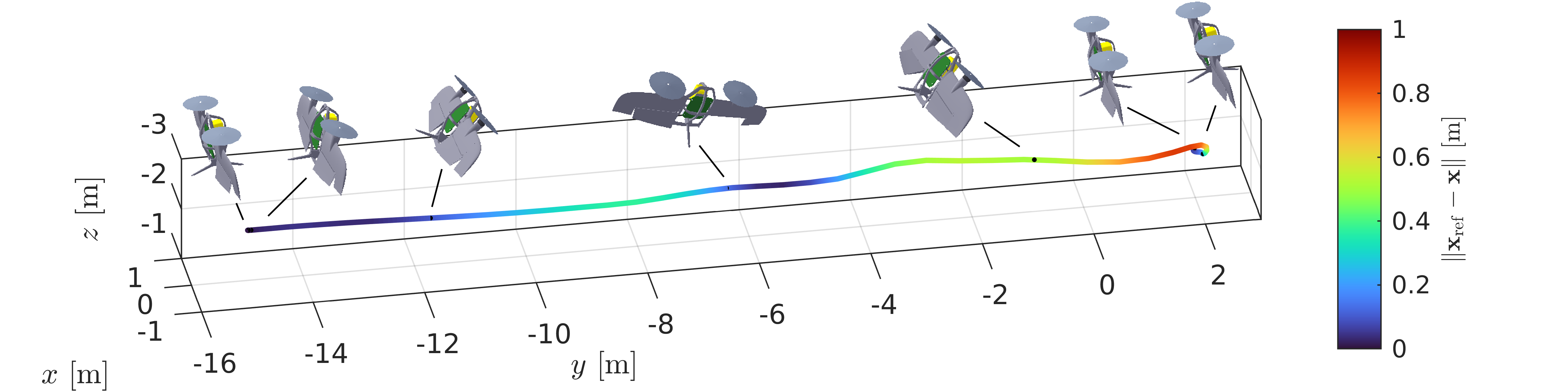}
		\caption{Experiment.}
		\label{fig:knifeedgetraj_exp}
	\end{subfigure}%
	\caption{Coordinated to Knife-Edge to Coordinated Flight. Interval between poses is 0.6 s.}
	\label{fig:knifeedgetraj}
\end{figure*}

Figure \ref{fig:knifeedgetraj} shows a straight trajectory between static hover start and end points.
The intermediate waypoints enforce a constant speed of 5 m/s and serve to transition between flight attitudes through the yaw reference $\psi_{\sref}$.
In the first of the three middle segments, the vehicle transitions from coordinated to knife-edge fight; in the second, it maintains constant knife-edge orientation; and in the third, it transitions back to coordinated flight.
Performing the transitions while maintaining straight flight at 5 m/s is challenging due to the aerodynamic interactions between vehicle attitude, flap deflections, and rotor speeds.
As expected, the position tracking error in the flight experiment increases at the transitions.
Once knife-edge orientation is reached, the position tracking error quickly reduces again.
The vehicle attitude during knife-edge flight differs somewhat between the reference and experiment trajectories.
The increased pitch angle in the experiment compensates for the neglected flap force contribution, and the small rotation towards the direction of travel compensates for the nonzero lateral force.
Finally, we note that the largest position tracking error is incurred close to the end point.
This error is mainly along the trajectory and is caused by delayed deceleration.
The maximum path error, \ie, position error with regard to the closest point on the trajectory line, occurs during the second transition and amounts to 47 cm.

\subsubsection{Climbing Turn}

\begin{figure*}
	\centering
	
	\begin{subfigure}[t]{0.5\textwidth}
		\centering
		\includegraphics[trim={0em 20em 0em 25em},clip,width=\linewidth]{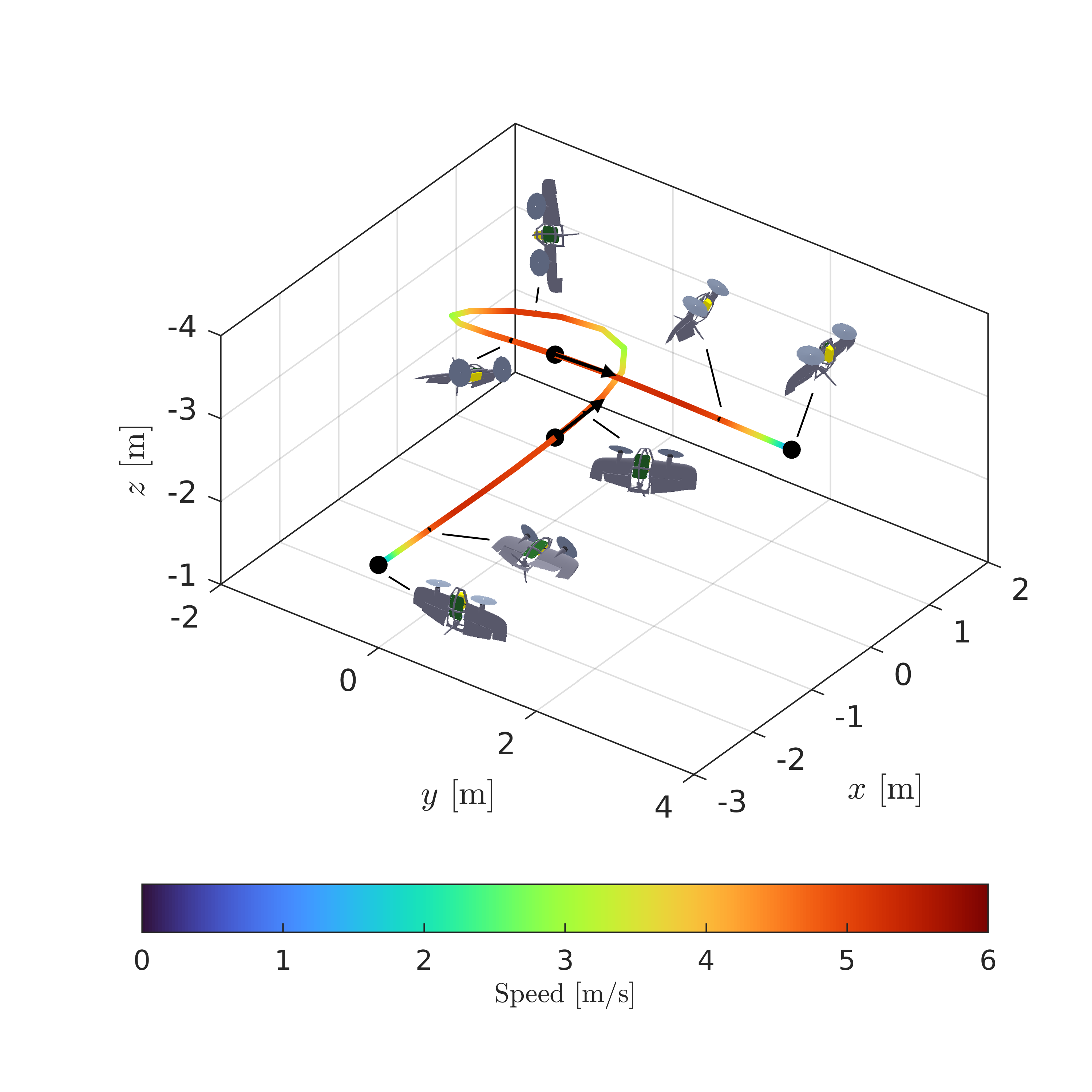}
		\caption{Reference with waypoints. Start and end points are static hover, and arrows indicate 5 m/s velocity constraints.}
		\label{fig:turn90_ref}
	\end{subfigure}%
	\begin{subfigure}[t]{0.5\textwidth}
		\centering
		\includegraphics[trim={0em 20em 0em 25em},clip,width=\linewidth]{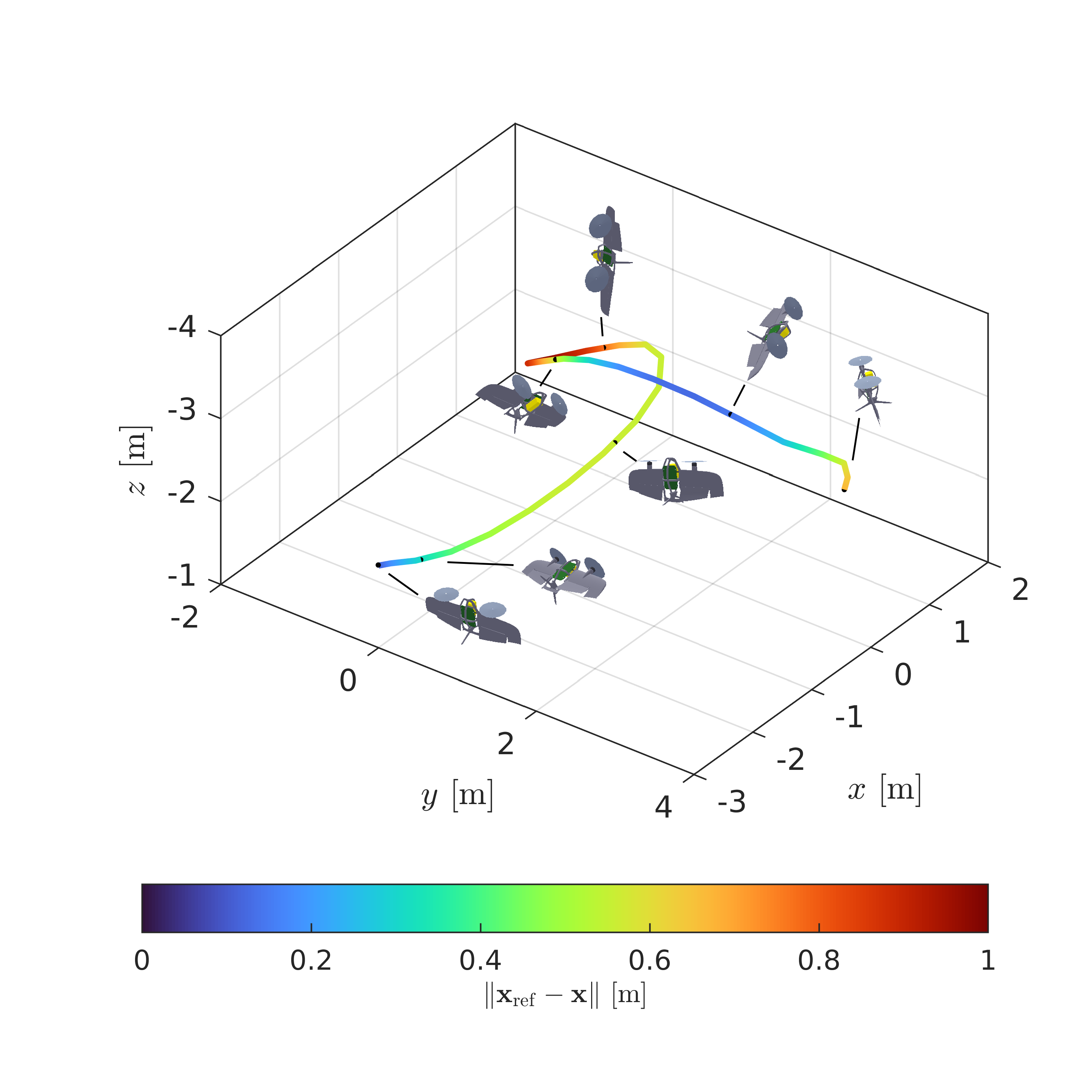}
		\caption{Experiment.}
		\label{fig:turn90_exp}
	\end{subfigure}%
	\caption{Climbing turn with 1 m height difference. Interval between poses is 0.5 s.}
	\label{fig:turn90}
\end{figure*}

We plan a climbing turn trajectory using four waypoints, as shown in Fig. \ref{fig:turn90}. The start and end points are constrained to static hover, and the two intermediate waypoints are positioned with only a height difference.
Using velocity constraints, we enforce straight and coordinated flight at these intermediate waypoints.
Hence, the entire 270 deg turn and 1 m climb occur between these two waypoints.
During the turn, the reference trajectory reaches about 90 banking angle, requires nearly the maximum motor speeds of 2500 rad/s, and reaches a peak angular rate of 11.3 rad/s (647 deg/s).
A peak load of 3.1$g$ is required during the turn, while the loads close to respectively the start and end points reach up to 2.0$g$. Consequently, the vehicle quickly completes the 11.3 m trajectory in 3.1 s, despite starting and ending in static hover.
In the flight experiment, we observe that, during the turn, a maximum load of 3.4$g$ and a peak angular rate of 10.9 rad/s (625 deg/s) are attained.
The motors briefly saturate, resulting in some loss of altitude. Once the saturation is resolved, the vehicle quickly catches up and reduces the position tracking error to below 20 cm before the turn is exited.

\subsubsection{Immelmann Turn}

\begin{figure*}
	\centering
	
	\begin{subfigure}[t]{0.5\textwidth}
		\centering
		\includegraphics[trim={65em 0em 30em 0},clip,width=\linewidth]{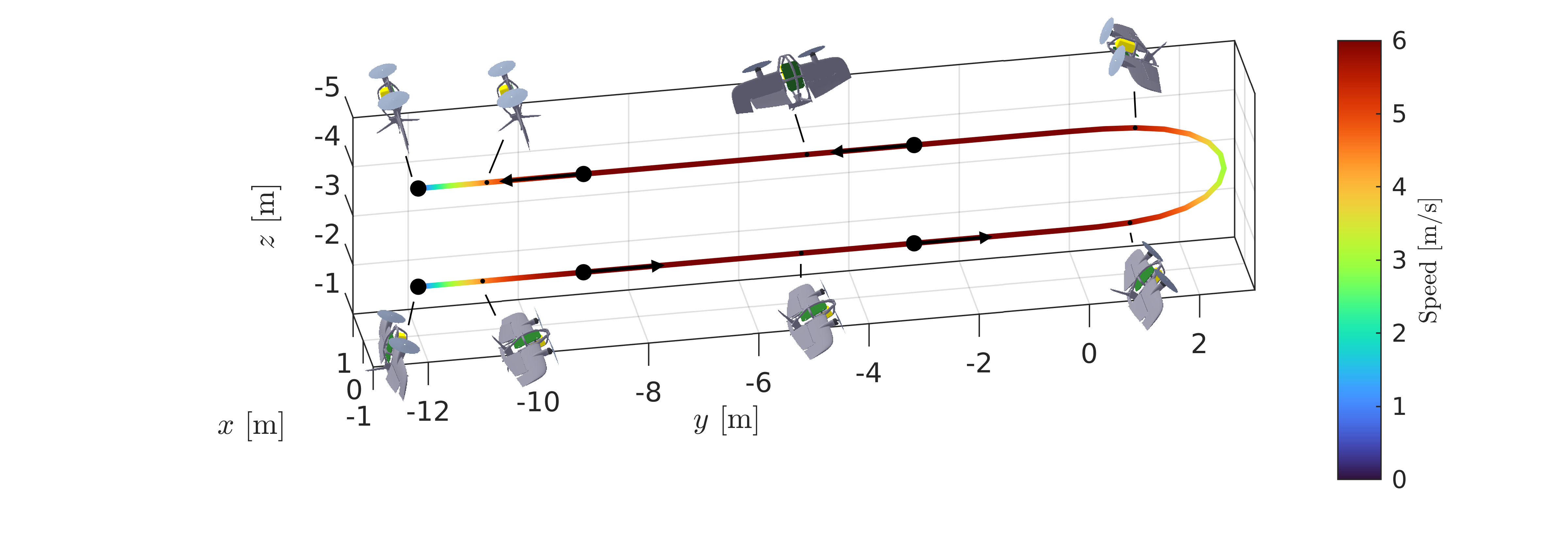}
		\caption{Reference with waypoints. Start and end points are static hover, and arrows indicate 6 m/s velocity constraints.}
		\label{fig:immel_ref}
	\end{subfigure}%
	\begin{subfigure}[t]{0.5\textwidth}
		\centering
		\includegraphics[trim={65em 0em 30em 0},clip,width=\linewidth]{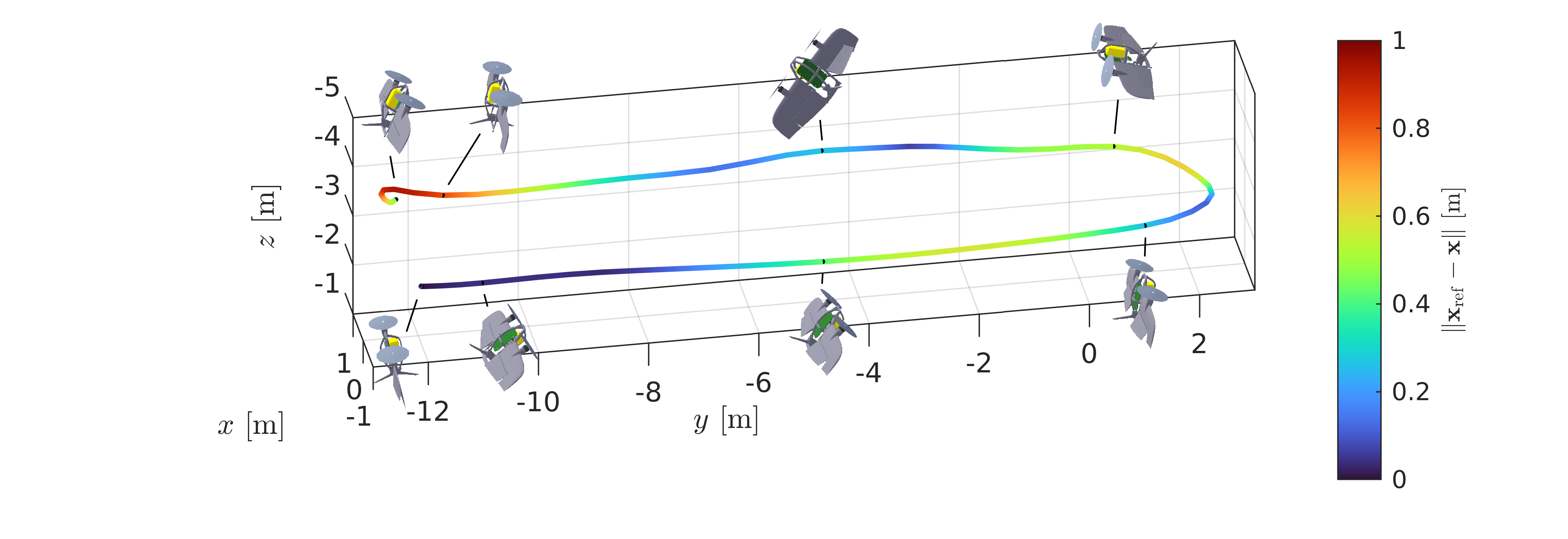}
		\caption{Experiment.}
		\label{fig:immel_exp}
	\end{subfigure}%
	\caption{Immelmann turn. Interval between poses is 1.0 s.}
	\label{fig:immel}
\end{figure*}

The Immelmann turn is a well-known aerobatics and aerial combat maneuver that reverses direction by performing a half loop followed by a half roll.
We generate the trajectory using static hover start and end points, and four intermediate waypoints.
The intermediate waypoints enforce constant speed coordinated flight prior to the half loop and constant speed transition from inverted to regular coordinated flight afterward.
Similar to the loop and knife-edge maneuvers described above, we observe increased error when exiting the loop segment, increased error during transition through uncoordinated flight orientation, and delayed deceleration towards the end point.
Comparison of the vehicle poses also leads to similar observations of small differences: increased pitch to account for flap force and increased yaw in uncoordinated flight to compensate for the nonzero lateral force.
The Immelmann turn combines several challenging aspects to exploit the expansive flight envelope of the tailsitter vehicle.
The maneuver contains large accelerations, inverted flight, and a transition through the entire yaw range (\ie, from $\psi_{\sref} = 0$ to $\psi_{\sref} = \pm \pi$ rad) at a peak angular rate of 9.4 rad/s (538 deg/s) while maintaining a linear speed of 6 m/s.
Based on snap minimization and differential flatness, the state-space trajectory and corresponding control inputs were generated efficiently and based on only four waypoints.
The flight experiment shows that the resulting maneuver can be tracked with acceptable position error ($< 60$ cm during the maneuver itself) while approaching the feasibility boundary, as over 90\% of the maximum flap deflection is reached during the half roll.

\subsubsection{Split S}

\begin{figure*}
	\centering
	
	\begin{subfigure}[t]{0.5\textwidth}
		\centering
		\includegraphics[trim={65em 0em 30em 0},clip,width=\linewidth]{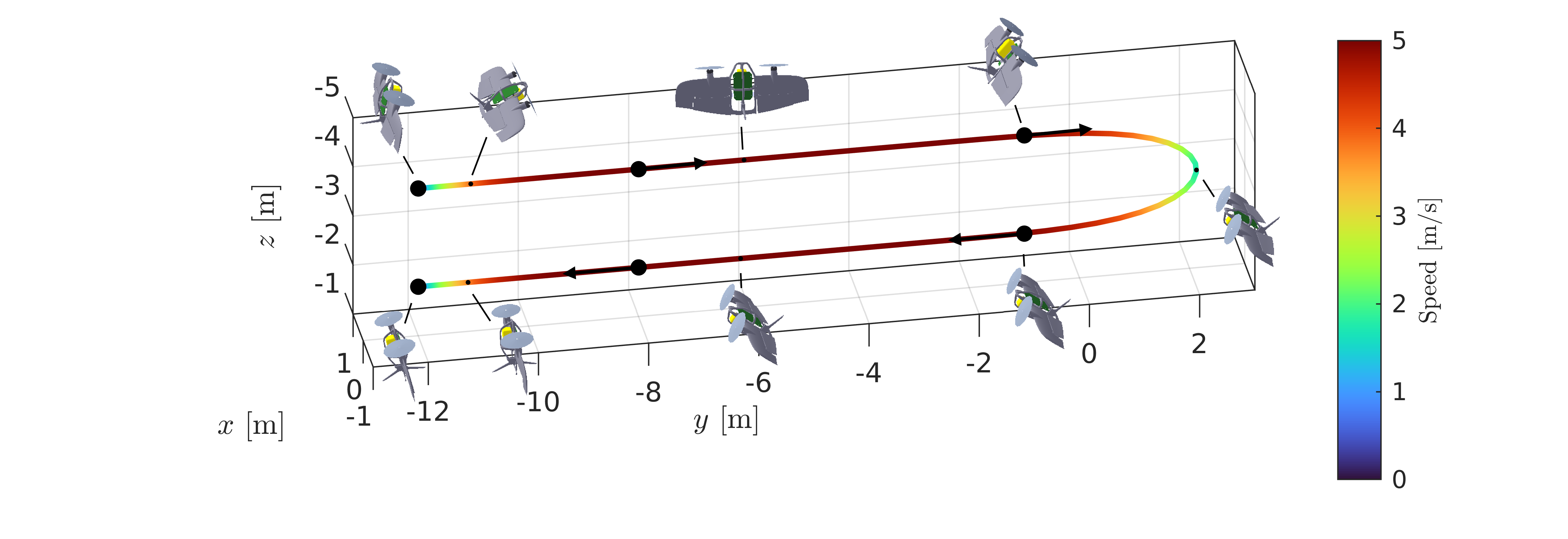}
		\caption{Reference with waypoints. Start and end points are static hover, and arrows indicate 5 m/s velocity constraints.}
		\label{fig:splits_ref}
	\end{subfigure}%
	\begin{subfigure}[t]{0.5\textwidth}
		\centering
		\includegraphics[trim={65em 0em 30em 0},clip,width=\linewidth]{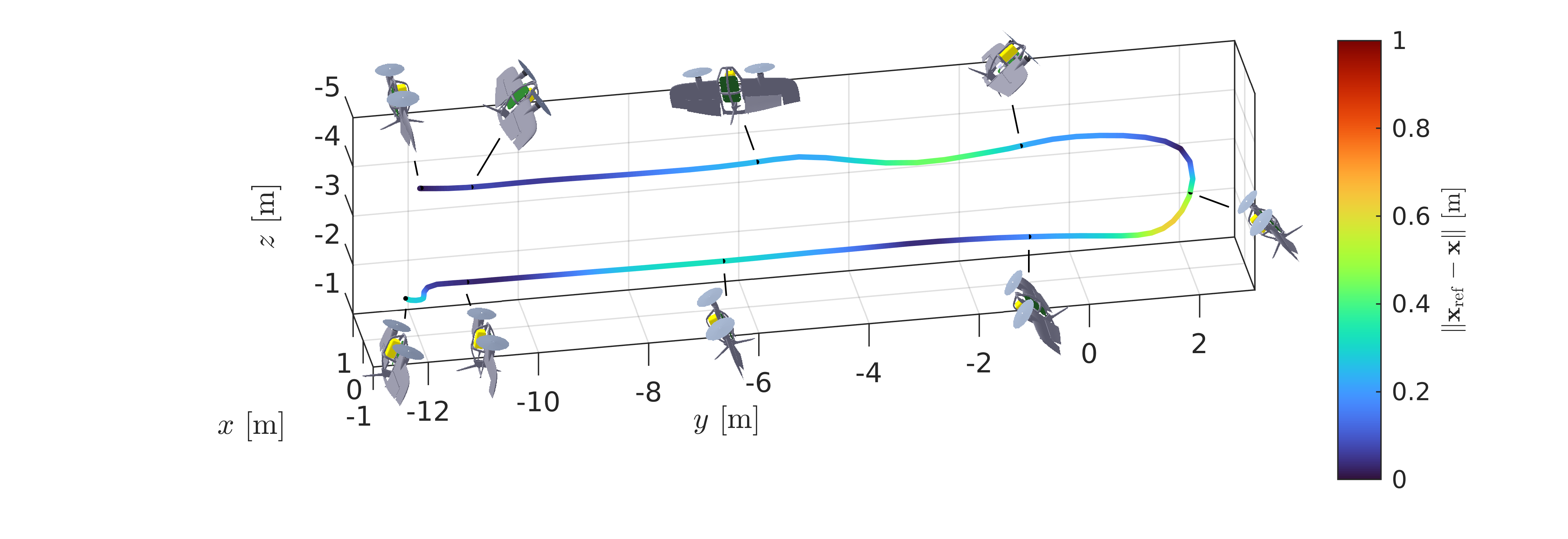}
		\caption{Experiment.}
		\label{fig:splits_exp}
	\end{subfigure}%
	\caption{Split S maneuver. Interval between poses is 1.0 s.}
	\label{fig:splits}
\end{figure*}

The Split S maneuver, shown in Fig. \ref{fig:splits}, is similar to the Immelmann but performed in opposite order.
The maneuver starts the top leg in coordinated flight, then transitions to inverted coordinated flight using the yaw reference $\psi_{\sref}$, and ends with a downward half loop that is exited in regular coordinated flight condition.
The trajectory is generated using similar waypoints as the Immelmann maneuver, albeit with opposite order and velocity direction.
Compared to the Immelmann turn, a smaller tracking error is achieved, because the flight speed is slightly lower (5 m/s versus 6 m/s) and because the trajectory ends with a relatively long stretch of coordinated flight, leading to a more stable deceleration.
We note the downward pitch motion during the half loop in both the reference and experiment trajectories.
By increasing the speed, we can obtain a more traditional Split S maneuver with a positive pitch rate. However, this maneuver requires a significantly larger flight volume.

\subsubsection{Differential Thrust Turn}

\begin{figure*}
	\centering
	
	\begin{subfigure}[t]{0.5\textwidth}
		\centering
		\includegraphics[trim={10em 0em 30em 0em},clip,width=\linewidth]{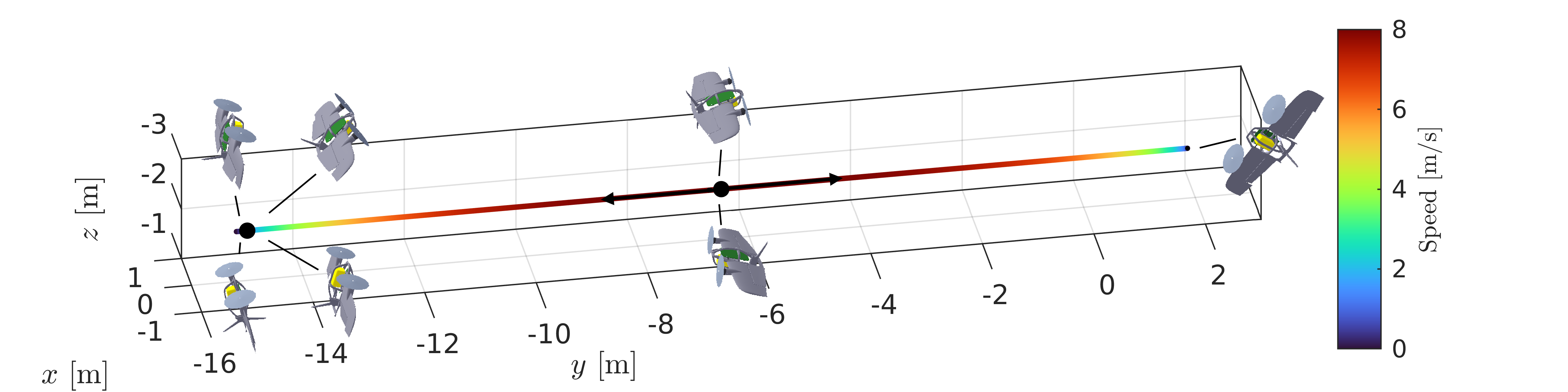}
		\caption{Reference with waypoints. Start and end points are static hover, and arrows indicate 8 m/s velocity constraints.}
		\label{fig:diffturn_ref}
	\end{subfigure}%
	\begin{subfigure}[t]{0.5\textwidth}
		\centering
		\includegraphics[trim={10em 0em 30em 0em},clip,width=\linewidth]{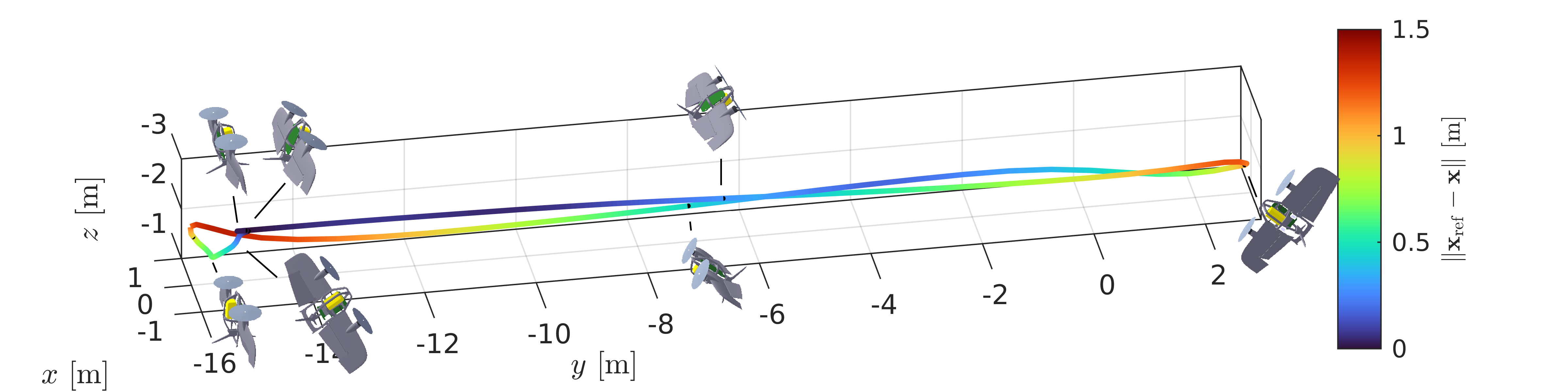}
		\caption{Experiment.}
		\label{fig:diffturn_exp}
	\end{subfigure}%
	\caption{Differential thrust turn. Interval between poses is 1.5 s.}
	\label{fig:diffturntraj}
\end{figure*}

The differential thrust turn, shown in Fig. \ref{fig:diffturntraj}, is an agile flight maneuver in which the vehicle reverses direction without deviating from a straight-line trajectory.
Unlike more traditional turns, which involve turning on a circular trajectory segment, the turn is performed by reorienting the vehicle using differential thrust and flap deflections, and then applying a large collective thrust to accelerate in the opposite direction.
The turn itself follows directly from snap minimization based on two coinciding waypoints with opposite velocity and yaw constraints.
In the flight experiment, a peak angular rate of 8.6 rad/s (493 deg/s) is reached during the turn.
As shown in the figure, the differential flatness transform is able to accurately predict the vehicle attitude at the midpoint of the turn.

\subsection{Racing Trajectory}\label{sec:trajts_mfbo}
\begin{figure*}
	\centering
	\begin{subfigure}[t]{0.49\textwidth}
		\centering
		\includegraphics[width=\linewidth]{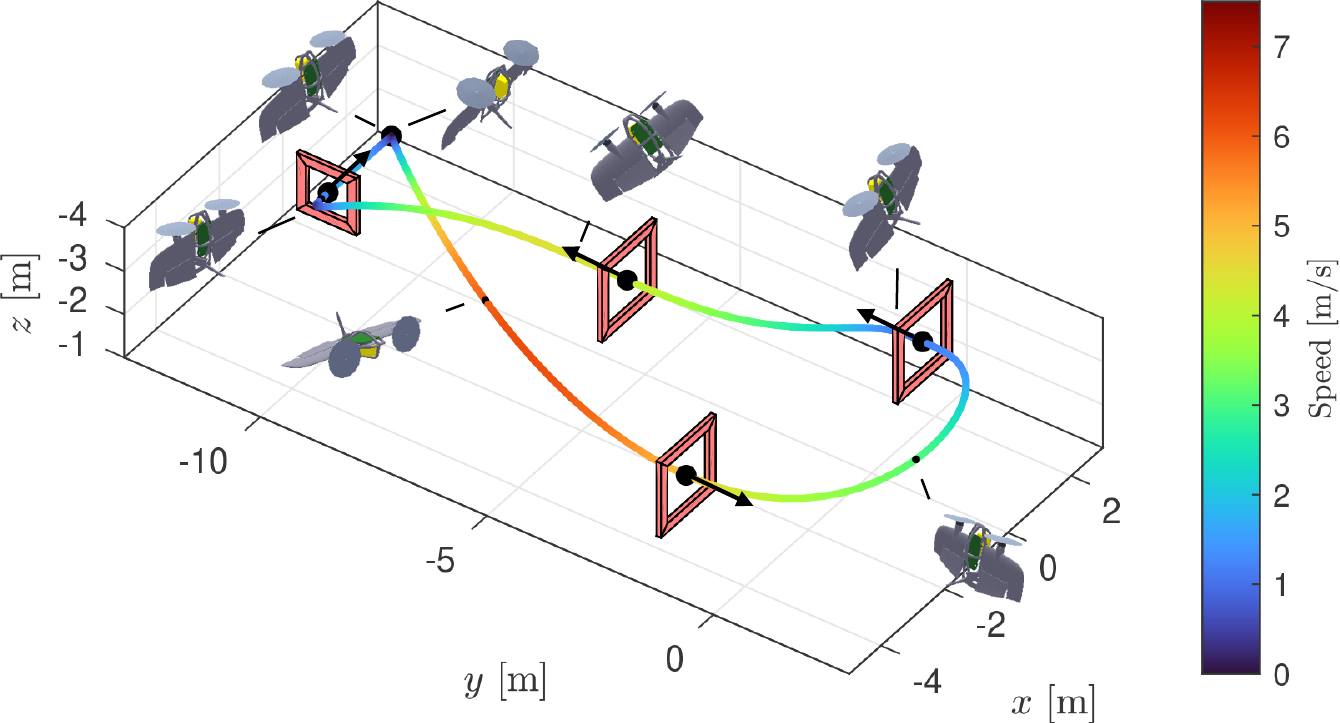}
		\caption{Using minimum-snap time allocation. Interval between poses is 2.3 s.}
		\label{fig:mfbo_gates_init}
	\end{subfigure}\hfill%
	\begin{subfigure}[t]{0.49\textwidth}
		\centering
		\includegraphics[width=\linewidth]{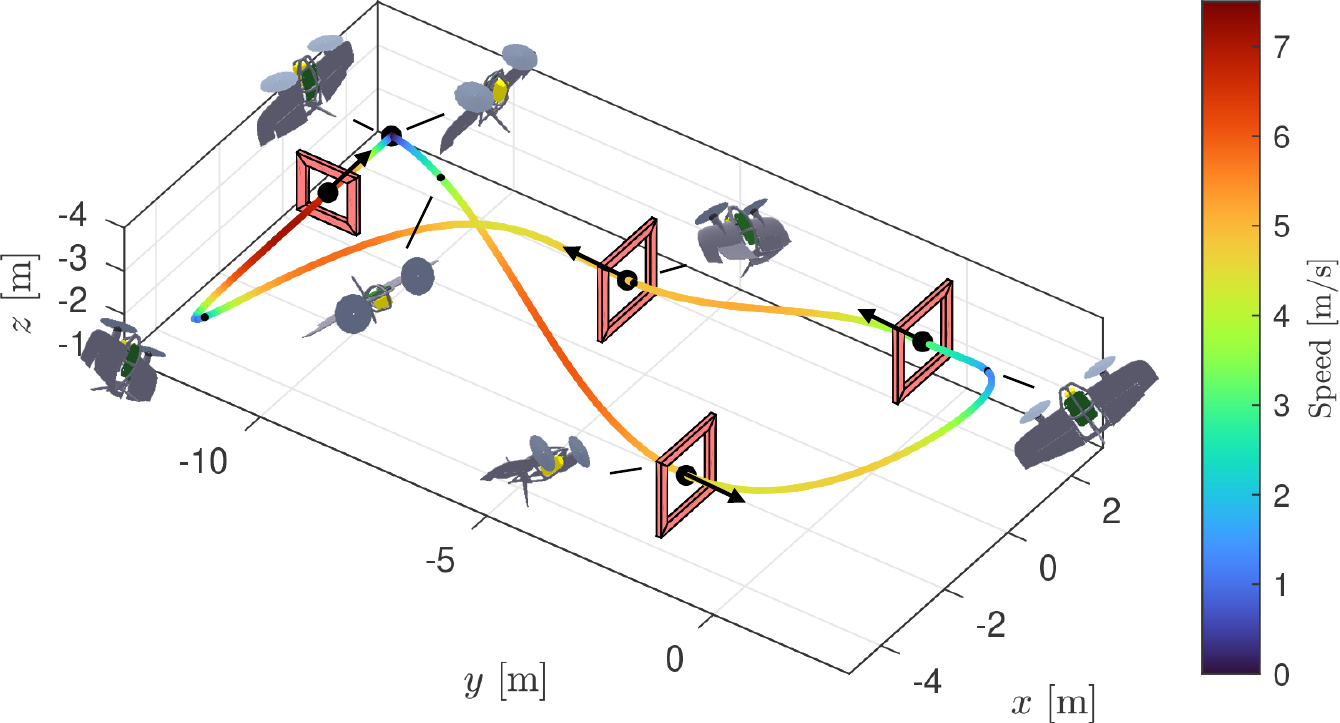}
		\caption{Using minimum-time time allocation. Interval between poses is 1.9 s.}
		\label{fig:mfbo_gates_final}
	\end{subfigure}%
	\caption{Minimum-snap racing trajectory through gates. Start and end points are static hover, and arrows indicate velocity direction constraints.}
	\label{fig:mfbo_gates}
\end{figure*}

In order to demonstrate agile high-speed flight in close proximity to obstacles, we generate a trajectory through a sequence of four drone racing gates.
The trajectory, shown in Fig. \ref{fig:mfbo_gates}, consists of six waypoints: coinciding start and end points constrained to static hover, and four gate waypoints with a directional velocity constraint that enforces flight perpendicular to the gate window.
Yaw is constrained so that the first three gates are passed in coordinated flight and the final, smaller gate in knife-edge flight, as it is too narrow to accommodate the tailsitter wingspan.

Instead of using the input constraint \eqref{eq:tstraj_feas1}, we scale $\vect{t}$ subject to the experimental feasibility constraint
\begin{equation}\label{eq:tstraj_feas2}
	\Sigma_T = \Big\{\vect{\sigma}_{\sref}\Big| \|\vect{x}_{\sref}(t) - \vect{x}(t)\|\leq 0.5\;\text{m}\;\;\;\;
	\forall t \in \left[0,T\right]\Big\},
\end{equation}
which guarantees that the vehicle does not collide with any of the gates.
The resulting trajectory is shown in Fig. \ref{fig:mfbo_gates_init}.
In order to obtain an even faster trajectory, we employ Bayesian optimization (BayesOpt) with experimental evaluations to further optimize $\vect{t}$~\cite{ryou2021multi}.
The BayesOpt algorithm, previously applied to quadrotors, can readily optimize the tailsitter trajectories by virtue of their flatness-based minimum-snap formulation.
The optimized time allocation $\vect{t}$ is then used to obtain a faster, more aggressive minimum-snap trajectory, shown in Fig. \ref{fig:mfbo_gates_final}.
This trajectory requires 19\% less flight time (11.1 s versus 13.7 s for the trajectory shown in Fig. \ref{fig:mfbo_gates_init}) but satisfies the same tracking accuracy constraint (\ie, \eqref{eq:tstraj_feas2}).
It has a maximum speed of 7.1 m/s.

The aggressive racing trajectory clearly shows how the tailsitter vehicle dynamics are exploited in trajectory generation to enable accurate tracking of fast and agile flight maneuvers.
In particular, during the knife-edge trajectory segments, differential thrust attitude control enables larger accelerations in the direction of flight.
The time-optimal trajectory exploits this additional acceleration to significantly increase the flight speed through the final gate.

\subsection{Aerobatic Sequence}
\begin{figure*}
	\centering
	
	\begin{subfigure}[t]{0.25\textwidth}
		\centering
		\includegraphics[trim={7em 0em 7em 0},clip,width=\linewidth]{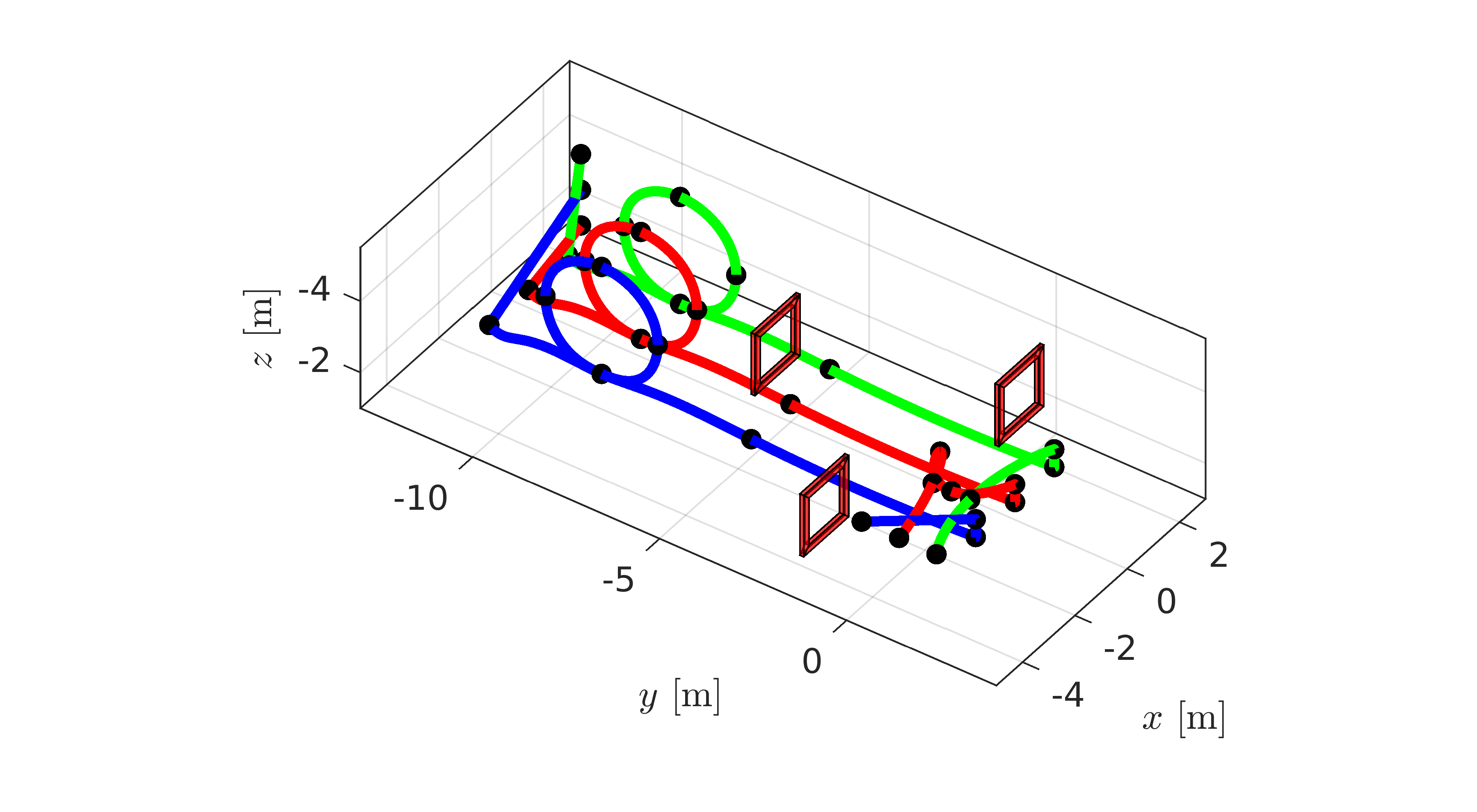}
		\caption{Transition and loops.}
		\label{fig:asp1}
	\end{subfigure}%
	\begin{subfigure}[t]{0.25\textwidth}
		\centering
		\includegraphics[trim={7em 0em 7em 0},clip,width=\linewidth]{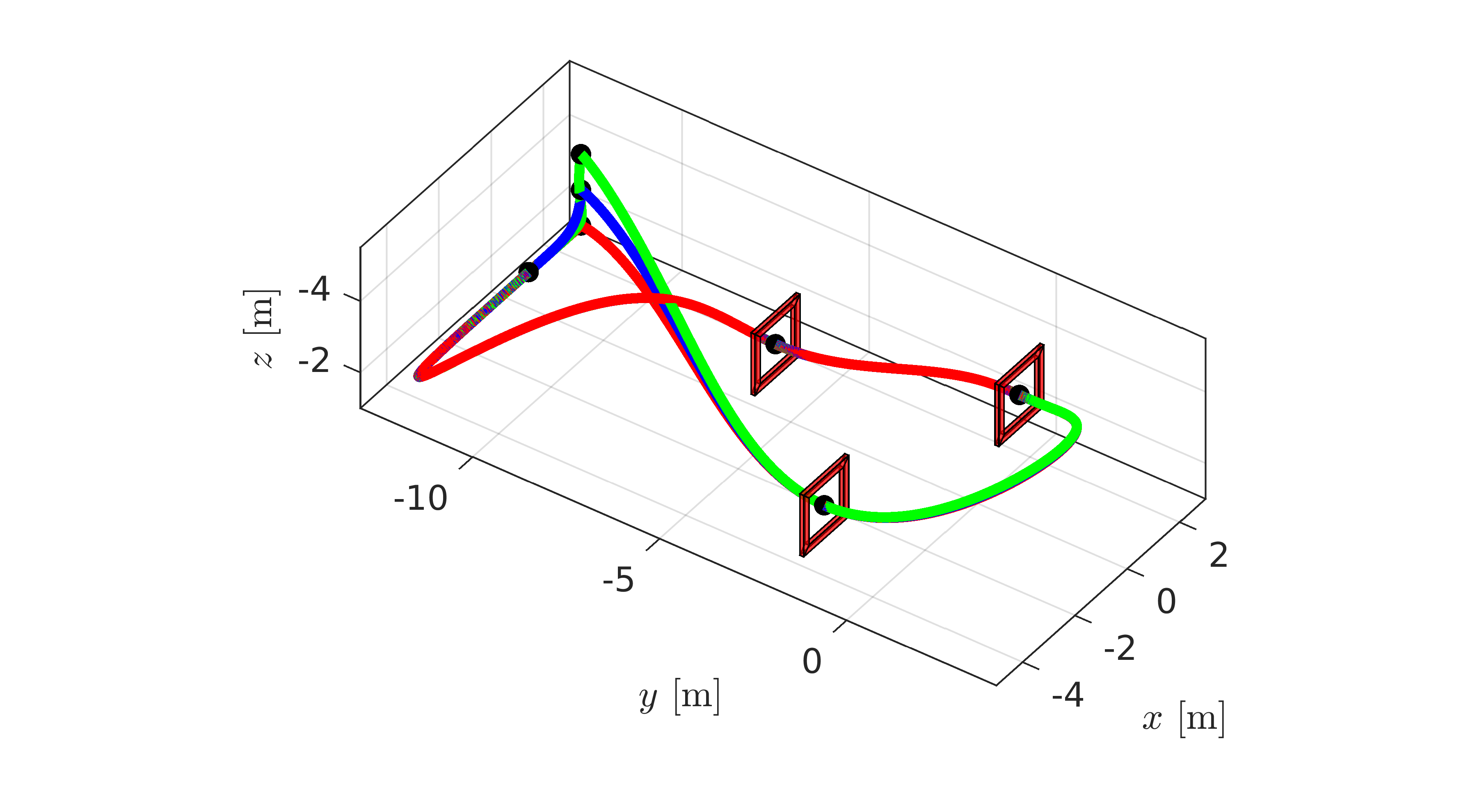}
		\caption{Close-proximity chase.}
		\label{fig:asp2}
	\end{subfigure}%
	\begin{subfigure}[t]{0.25\textwidth}
		\centering
		\includegraphics[trim={7em 0em 7em 0},clip,width=\linewidth]{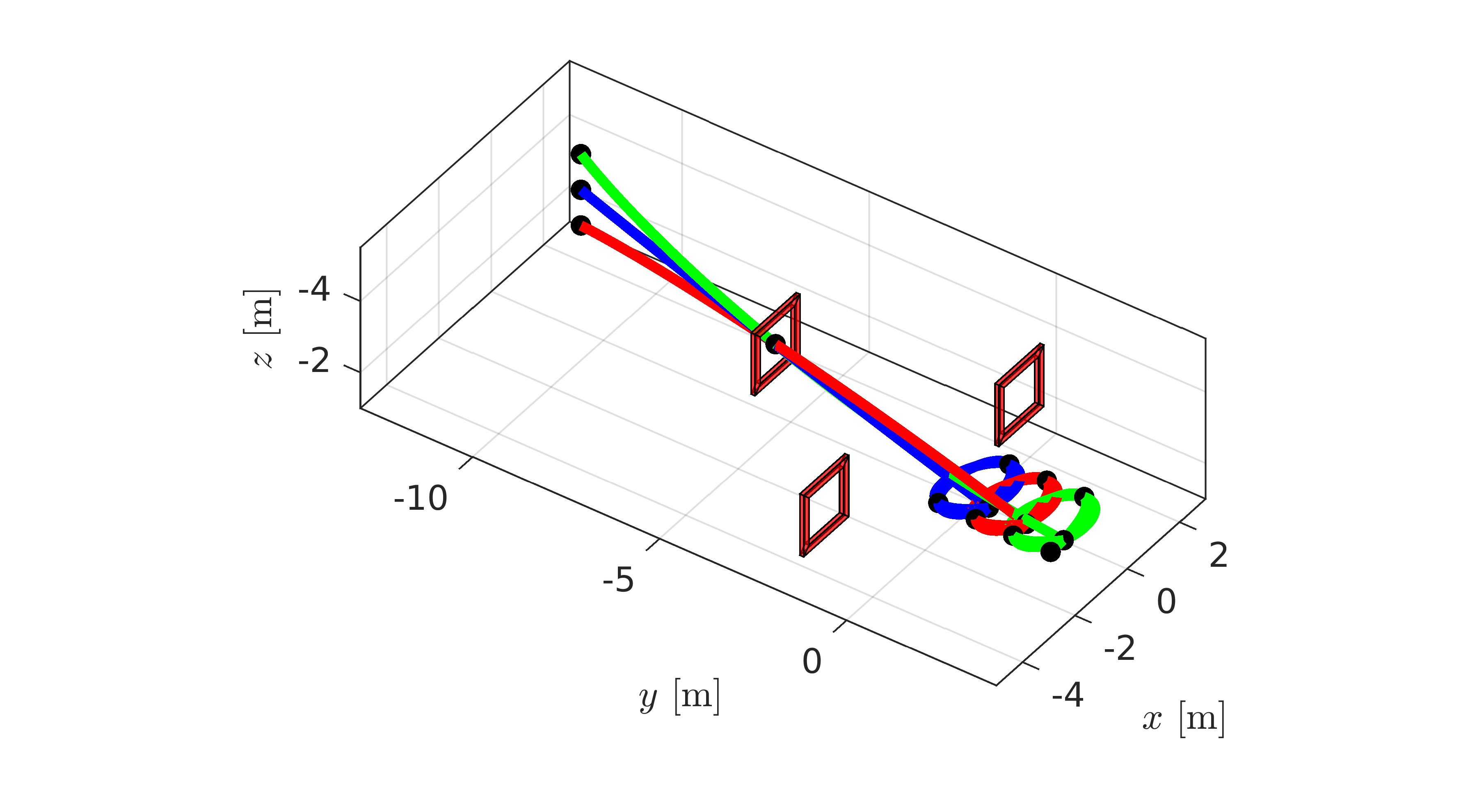}
		\caption{Knife edge and hover.}
		\label{fig:asp3}
	\end{subfigure}%
	\begin{subfigure}[t]{0.25\textwidth}
		\centering
		\includegraphics[trim={7em 0em 7em 0},clip,width=\linewidth]{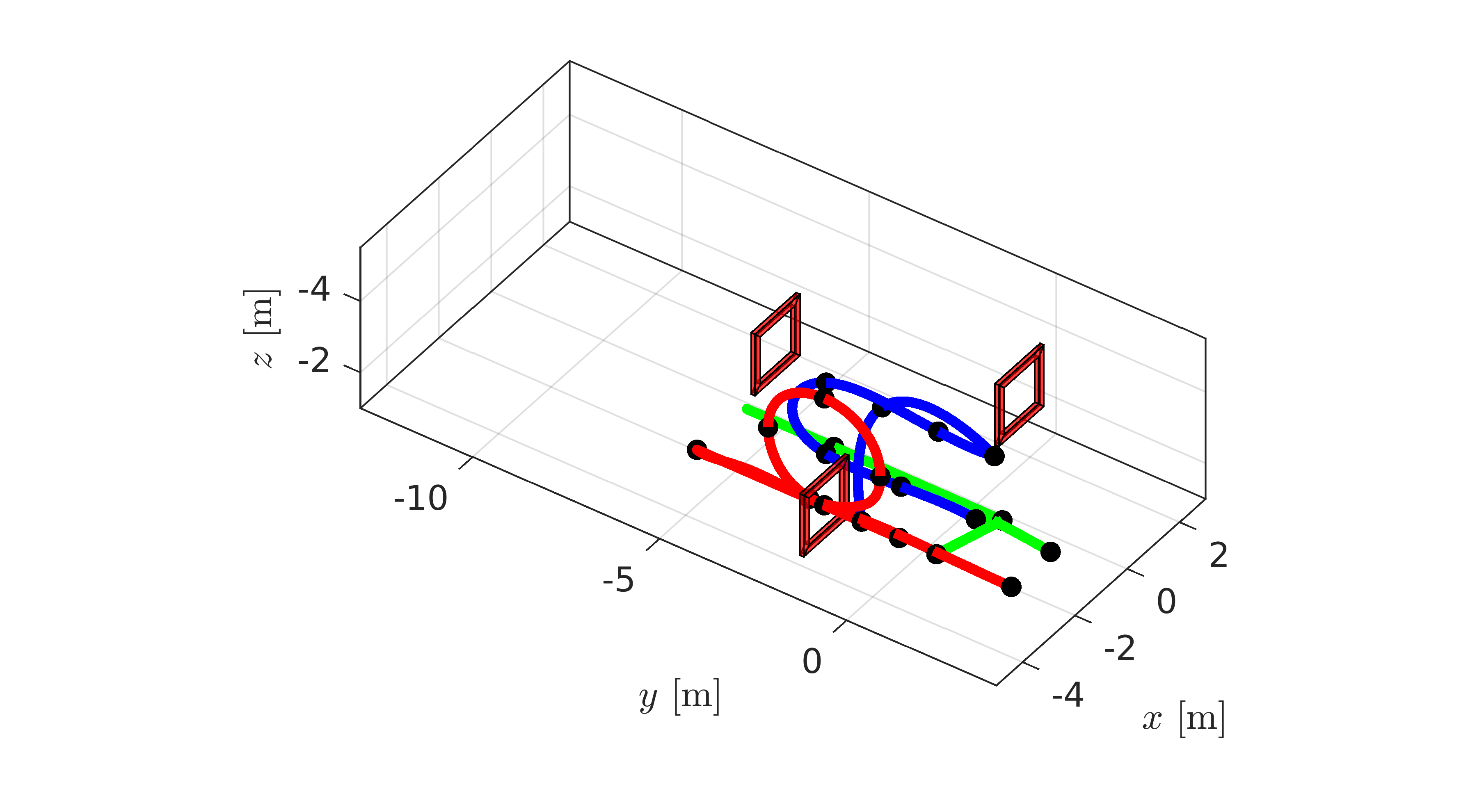}
		\caption{Loop and aerobatic turns.}
		\label{fig:asp4}
	\end{subfigure}%
	\caption{Multi-vehicle aerobatic sequence for three tailsitter aircraft.}
	\label{fig:as}
\end{figure*}

As a final demonstration of the consistency and accuracy with which the generated aerobatic trajectories can be flown, we perform an airshow-like aerobatic sequence with three tailsitter aircraft.
The sequence, shown in the accompanying video and Fig. \ref{fig:as}, consists of four stages that seamlessly follow each other and incorporate many of the aerobatic maneuvers described in Section \ref{sec:aerobatic} as well as the racing trajectory described in Section \ref{sec:trajts_mfbo}.
During the first stage, the three vehicles synchronously transition from hover to coordinated flight at 5.8 m/s and perform a loop.
In the second stage, the tailsitters fly the minimum-time racing trajectory through the gates with only 0.7 s separation between successive vehicles.
The third stage starts with successive transitioning flight from coordinated to knife-edge condition through the center gate, which is followed by synchronous aggressive maneuvers in hover with horizontal accelerations up to 11.5 m/s\textsuperscript{2} while maintaining 45 cm separation between adjacent vehicles.
Finally, the three vehicles synchronously perform respectively the Immelmann turn, the differential thrust turn, and a loop through one of the gates.

\section{Conclusion}\label{sec:conclusion}
We proposed the novel application of snap minimization towards aerobatic trajectory generation for a tailsitter flying wing.
The method plans trajectories in the flat output space, instead of considering computationally expensive optimization on the more complicated state and control input space.
Through experimental validation, it was shown that the derived flatness transform provides a useful prediction of the critical trajectory time or speed at which a stark increase in tracking error occurs on the real vehicle.
The proposed algorithm was used to generate trajectories for six aerobatic maneuvers, a race course through several gates, and an airshow-like aerobatic sequence for three tailsitters.
We found that the real vehicle was indeed capable of accurately tracking these aggressive trajectories and that the vehicle pose and control inputs predicted by the flatness transform closely matched those of the actual vehicle.
In conclusion, the proposed algorithm accurately plans aerobatic trajectories that exploit the expansive flight envelope of the tailsitter flying wing, without requiring costly optimization on the state and control input space.
\section*{Acknowledgments}
We thank Murat Bronz and John Aleman for the design, fabrication, and assembly of the aircraft used in the experiments.
This work was supported by the Army Research Office through grant W911NF1910322.

\bibliographystyle{IEEEtran}
\bibliography{refs}

\end{document}